\documentclass[12pt]{article}
\usepackage{amsmath}
\usepackage{times}
\usepackage{graphicx}
\usepackage{color}
\usepackage{multirow}
\usepackage[authoryear]{natbib}

\usepackage{rotating}
\usepackage{bbm}
\usepackage{latexsym}

\newcommand{\captionfonts}{\normalsize}

\makeatletter  
\long\def\@makecaption#1#2{%
  \vskip\abovecaptionskip
  \sbox\@tempboxa{{\captionfonts #1: #2}}%
  \ifdim \wd\@tempboxa >\hsize
    {\captionfonts #1: #2\par}
  \else
    \hbox to\hsize{\hfil\box\@tempboxa\hfil}%
  \fi
  \vskip\belowcaptionskip}
\makeatother

\usepackage[latin1]{inputenc} 
\usepackage[english]{babel} 
\usepackage{graphicx,xcolor} 
\usepackage{amssymb,amsthm,mathtools} 
\usepackage{hyperref,url} 
\usepackage{soul} 
\usepackage{enumitem}

\usepackage[linesnumbered,ruled,vlined]{algorithm2e}
\SetKwInput{KwInput}{Input}   
\SetKwInput{KwOutput}{Output} 
\makeatletter
\renewcommand*{\@algocf@pre@ruled}{}
\renewcommand*{\@algocf@post@ruled}{}
\makeatother

\usepackage{comment} 
\usepackage{cleveref}

\newtheorem{proposition}{Proposition}
\newtheorem{lemma}{Lemma}
\newtheorem{theorem}{Theorem}
\newtheorem*{remark}{Remark}

\usepackage{bm} 

\def\rmM{\mathrm{M}}
\def\rmP{\mathrm{P}}

\def\NMF{\texttt{NMF}}

\def\SON{\textrm{SON}}

\usepackage{bm}                 
\DeclareMathOperator*{\argmin}{\textrm{argmin}\,} 
\def\cone            {\textrm{cone}}     
     
\def\ones            {\bm{1}}            
\def\proj            {\textrm{proj}}     
\def\rk              {\textrm{rank}}     
\def\st              {\,\textrm{s.t.}\,} 
\def\zeros           {\bm{0}}            
\def\IN {\mathbb{N}} 		
\def\IR  {\mathbb{R}} 	    
\def\barIR{\overline{\IR}}  
\def\IRm {\IR^{m}} 			
\def\IRn {\IR^{n}} 			
\def\IRmn{\IR^{m \times n}} 
\def\IRmr{\IR^{m \times r}} 
\def\IRrn{\IR^{r \times n}} 

\def\A {\bm{A}}

\def\D {\bm{D}}

\def\G {\bm{G}}
\def\H {\bm{H}} 
\def\I {\bm{I}}

\def\M {\bm{M}}
\def\N {\bm{N}}

\def\Q {\bm{Q}}
\def\R {\bm{R}}

\def\U {\bm{U}}
\def\V {\bm{V}}
\def\W {\bm{W}}
\def\X {\bm{X}} 

\def\Z {\bm{Z}}

\def\b {\bm{b}}
\def\c {\bm{c}}

\def\h {\bm{h}}

\def\m {\bm{m}}

\def\u {\bm{u}}
\def\v {\bm{v}}
\def\w {\bm{w}}
\def\x {\bm{x}}

\def\cN { \mathcal{N}}
\def\cO { \mathcal{O}}

\def\bLambda  {\boldsymbol{\Lambda}}

\def\bSigma   {\boldsymbol{\Sigma}}

\def\bxi      {\boldsymbol{\xi}}

\usepackage{tikz}
\usetikzlibrary{graphs,graphs.standard}
\usepackage{pgfplots,pgfplotstable}

\usepackage{authblk}
\author[1]{$^1$Andersen Ang
~~~~
Waqas Bin Hamed
~~~~
$^2$Hans De Sterck
\\
\normalsize $^1$School of Electronics and Computer Science, University of Southampton, UK
\\
$^2$Department of Applied Mathematics, University of Waterloo, Canada
}

\title{\vspace{-17.5mm}
\LARGE 
Sum-of-norms regularized \\ Nonnegative Matrix Factorization
\thanks{
Andersen Ang (\url{andersen.ang@soton.ac.uk}) is the corresponding author.
Part of the work of this paper was done when Andersen Ang was a post-doctoral fellow and when Waqas Bin Hamed was a master student, both at the University of Waterloo.
Funding: Andersen Ang acknowledge the supported in part by a joint postdoctoral fellowship by the Fields Institute for Research in Mathematical Sciences and the University of Waterloo, and in part by Discovery Grants from the Natural Sciences and Engineering Research Council (NSERC) of Canada.
}
}
\date{\today}
\begin{document}
\maketitle

\begin{abstract}
When applying nonnegative matrix factorization (NMF), the rank parameter is generally unknown.
This rank, called the nonnegative rank, is usually estimated heuristically since computing its exact value is NP-hard.
In this work, we propose an approximation method to estimate the rank on-the-fly while solving NMF.
We use the sum-of-norm (SON), a group-lasso structure that encourages pairwise similarity, to reduce the rank of a factor matrix when the initial rank is overestimated.
On various datasets, SON-NMF can reveal the correct nonnegative rank of the data without prior knowledge or parameter tuning.

SON-NMF is a nonconvex, nonsmooth, non-separable, and non-proximable problem, making it nontrivial to solve.
First, since rank estimation in NMF is NP-hard, the proposed approach does not benefit from lower computational complexity.
Using a graph-theoretic argument, we prove that the complexity of SON-NMF is essentially irreducible.
Second, the per-iteration cost of algorithms for SON-NMF can be high. 
This motivates us to propose a first-order BCD algorithm that approximately solves SON-NMF with low per-iteration cost via the proximal average operator.

SON-NMF exhibits favorable features for applications.
Besides the ability to automatically estimate the rank from data, SON-NMF can handle rank-deficient data matrices and detect weak components with small energy.
Furthermore, in hyperspectral imaging, SON-NMF naturally addresses the issue of spectral variability.

\vspace{2mm}
\end{abstract}

\section{Introduction}\label{sec:intro}
\paragraph{Nonnegative Matrix Factorization (NMF)} 
Denote \NMF($\M,r$) as the following problem: 
given a matrix $\M \in \IRmn_+$, find two factor matrices $\W \in \IRmr_+$ and $\H \in \IRrn_+$ such that $\M = \W\H$.
NMF \cite{paatero1994positive, gillis2020nonnegative} describes a cone: $\M$ is a point cloud (of $n$ points) in $\IRm_+$, contained in a polyhedral cone generated by the $r$ columns of $\W$, with nonnegative weights encoded in $\H$. 
Here, $H_{ij}$ represents the contribution of column $\w_i$ to the representation of data column $\m_j$; see, e.g., \cite[Fig.1]{leplat2019minimum}.

\paragraph{Nonnegative rank} 
Let $r = \rk_+(\M)$ denote the nonnegative rank of a matrix, where $r$ is the minimal number of nonnegative rank-1 components required to represent $\M$
\cite[Sect.4]{berman1994nonnegative}, \cite[Sect.3]{gillis2020nonnegative},  i.e.,
\begin{equation}\label{NMF:rank:eqvi}
\M 
= \W \H 
=
\begin{bmatrix}
\w_1 \dots \w_r    
\end{bmatrix}
\begin{bmatrix}
    \h^1  \\ \vdots \\ \h^r
\end{bmatrix}
= \w_1\h^1 + \dots + \w_r \h^r
= \sum_{\ell=1}^r  \w_{\ell} \h^{\ell},
\tag{$\NMF(\M,r)$}
\end{equation}
where $\w_j \geq \zeros$ is the $j$th column of $\W$, and $\h^j \geq \zeros$ is the $j$th row of $\H$.
Here, $\w_j\h^j$ represents the $j$th rank-1 factor in $\W\H$.

\paragraph{$r$ is important}
The parameter $r$ controls the model complexity of NMF and plays a critical role in data analysis.
In signal processing \cite{leplat2020blind}, $r$ represents the number of sources in an audio signal.
If $r$ is overestimated, overfitting occurs, where the extra components in the model capture noise (e.g., piano mechanical noise \cite[Sect.4.2]{ang2020nonnegative}) rather than meaningful information. 

\paragraph{$r$ is unknown}
Generally, $r$ is unknown.
Finding $r$ in \ref{NMF:rank:eqvi} for $\rk_+(\M) \geq 3$ is NP-hard \cite{vavasis2010complexity}\footnote{Note that $\rk_+(\M)$ is not the same as $\rk(\M)$, which can be computed by eigendecomposition or singular value decomposition. 
See \cite{gillis2020nonnegative} for solving \NMF($\M,r$) in the case $\rk_+(\M) \leq 2$.}.
In many cases, $\rk(\M)$ and/or $\rk_+(\M)$ are small since $\M$ is approximately low-rank \cite{udell2019big} and/or has low nonnegative rank \cite[Sect.9.2]{gillis2020nonnegative}.
Heuristics have been proposed to find $r$. 
Besides trial-and-error, the two main groups of methods for estimating $r$ are stochastic/information-theoretic and algebraic/deterministic.
The first group includes Bayesian methods \cite{tan2012automatic}, the cophenetic correlation coefficient \cite{esposito2020nmf}, and minimum description length \cite{squires2017rank}.
The second group includes fooling sets \cite{cohen1993nonnegative} and the $f$-vector in combinatorics \cite{dewez2021geometric}.
See \cite[Sect.3]{gillis2020nonnegative} for a summary of the algebra of $\rk_+$.

In this work, we focus on approximately solving \ref{NMF:rank:eqvi} without knowing $r$ in advance.
This is achieved by imposing a ``rank penalty'' on NMF.
Instead of using the nuclear norm nor the rank itself as a penalty term, we consider a clustering regularizer called Sum-of-norms (SON): we propose SON-NMF to ``relax'' the assumption of knowing $r$.
Before we introduce SON-NMF, we first review the SON term.

\paragraph{Matrix $\ell_{p,q}$-norm} 
The $\ell_{p,q}$-norm of a matrix $\X \in \IRmn$ is defined as
\[
\| \X \|_{p,q} 
~\coloneqq~
\left(
\sum_{j=1}^n
\Bigg(
\sqrt[p]{\sum_{i=1}^m X_{ij}^p}
\Bigg)^q
\right)^{\frac{1}{q}}
~=~
\left\| 
\begin{bmatrix}
\| \x_1 \|_p
\\
\vdots
\\
\| \x_n \|_p
\end{bmatrix}
\right\|_q,
\]
where, in the last equality, we first take the $p$-norm of each column and then take the $q$-norm of the resulting vector.
A popular choice of the $\ell_{p,q}$-norm is the $\ell_{2,1}$-norm, which is widely used in the multiple measurement vector problem \cite{cotter2005sparse}, sparse coding \cite{nie2010efficient}, and robust NMF \cite{kong2011robust}.

\paragraph{Sum-of-norms (SON)}
We define the SON of a matrix $\X$ as the $\ell_{2,1}$-norm of $P(\X)$, where $\X \mapsto P(\X)$ is all the pairwise difference $\x_i - \x_j$.
As $\| \x_i - \x_j\|_2 = \| \x_j - \x_i \|_2$, there are $\frac{n^2-n}{2}$ terms in SON of $\X$.
In this work, we propose using SON$_{2,1}(\W)$ as a regularizer for NMF, to be presented in the next section.
Below, we give remarks on SON$_{2,q}(\W)$ for other choices of $q$.
\begin{itemize}[leftmargin=12pt]
\item SON$_{2,0}(\W)$ with $q=0$: It is trivial that $\rk(\W) \leq \SON_{2,0}(\W)$, because the set of linearly independent vectors is a subset of the set of unequal vector pairs.
By the combinatorial nature of the $\ell_0$-norm, minimizing $\SON_{2,0}(\W)$ is NP-hard, and its complexity scales with $r$. 
Therefore, SON$_{2,0}(\W)$ is computationally unfavourable for NMF applications with large $r \approx (m,n)$, which is the case in this work.

\item SON$_{2,2}(\W)$ with $q=2$: By definition, this is the Frobenius norm of $P(\W)$.
This SON has been used in graph-regularized NMF \cite{cai2010graph}, but it differs from \eqref{eqn:sonnmf} for two reasons: (1) the graph regularizer is a weighted squared $\SON_{2,2}$ norm, which is everywhere differentiable, unlike SON$_{2,1}(\W)$; and (2) SON$_{2,2}(\W)$ does not induce sparsity, whereas SON$_{2,1}(\W)$ does.

\item SON$_{2,\infty}(\W)$ with $q \rightarrow \infty$: This term focuses on the pair $(\w_i,\w_j)$ that is mutually furthest apart, ignoring the rest.
This is unfavourable for removing redundant $\w_j$ in NMF for the purposes of this work.
\end{itemize}

We are now ready to introduce SON-NMF.

\paragraph{SON-NMF} 
We propose to regularize NMF by 
$\textrm{SON}_{2,1}(\W) 
= \sum_{ i \neq j }\| \w_i - \w_j \|_2
$ as
\begin{equation}\label{eqn:sonnmf}
\begin{array}{rl}
\displaystyle
\argmin_{\W,\H} F(\W,\H) 
\coloneqq 
&\displaystyle
\dfrac{1}{2}\| \W\H - \M \|_F^2 
+ \lambda \sum_{i \neq j} \| \w_i - \w_j \|_2
\\
& \displaystyle
+ \gamma \sum_{i} \| \max\{ -\w_i , \zeros\}\|_1 
+ \iota_{\Delta^r}(\H),
\end{array}
\tag{SON-NMF}
\end{equation}
where $\frac{1}{2} \| \M - \W\H \|_F^2 : \IRmn \times \IRmr \times \IRrn  \rightarrow \IR$ is a smooth, nonconvex data-fitting term,
the constants $\lambda >0$  and $\gamma>0$ are parameters,
the functions $\sum_{i} \| \max\{ -\w_i , \zeros\}\|_1$  and $\iota_{\Delta^r}(\H) = \sum_j \iota_{\Delta^r}(\h_j)$  
are nonsmooth, lower-semicontinuous, proper convex functions representing model constraints: respectively, 
the nonnegativity of $\w_j$ (i.e., $\W \geq \zeros$) and the requirement that $\h_j$ lies in the $r$-dimensional unit simplex (i.e., $\H$ is element-wise nonnegative and $\H^\top \ones_r \leq \ones_n$, where $\ones_r \in \IR^r$ denotes a vector of ones). 
Note that in \eqref{eqn:sonnmf} we use the penalty $\sum_{i} \| \max\{ -\w_i , \zeros\}\|_1$, which enforces   nonnegativity  $\W \geq \zeros$ for sufficiently large $\lambda$, this will be explained in \cref{sec:algo21W}. 
We defer the definition of symbols used in \eqref{eqn:sonnmf} to the end of this section.

\paragraph{SON encourages multicollinearity and rank-deficiency for NMF}\label{remark:multicollinear}
The SON term encourages the pairwise difference in $\|\w_i - \w_j\|_2$ to be small,  potentially resulting in multicollinearity in the matrix $\W$.
Note that in traditional regression models, multicollinearity is strongly discouraged due to its negative statistical effects on the variables \cite{farrar1967multicollinearity}.
In this work, we intentionally promote multicollinearity in $\W$ to encourage rank deficiency, which helps reduce an overestimated rank during rank estimation.
In other words, SON-NMF can be seen as the ordinary NMF model with a multicollinearity regularizer: the rank of $\W$ is overestimated at the first iteration, and the regularizer gradually reduces it to the correct value during the algorithmic process.

There is a ``price to pay'' for such multicollinearity.
If $\W$ is near-multicollinear, its condition number is large, making $\W^\top \W$ ill-conditioned and negatively affecting the process of updating $\H$.
See the discussion in Section~\ref{sec:algo21}.

\paragraph{Contributions}
We introduce a new problem \eqref{eqn:sonnmf} with the contributions:
\begin{itemize}[leftmargin=12pt]
\item \textbf{Empirically rank-revealing.}
On synthetic and real-world datasets, we empirically show that model~\eqref{eqn:sonnmf}, free from tuning the rank $r$, will itself find the correct $r$ in the data automatically when $r$ is overestimated.
This is due to the sparsity-inducing property of the $\ell_{2,1}$ norm in SON$_{2,1}$.
\begin{itemize}[leftmargin=10pt]
\item \textbf{Rank-deficient compatibility.}
SON-NMF can handle rank-deficient problems, i.e., data matrices whose true rank is smaller than the overestimated parameter $r$.
This has two advantages.
First, it prevents overfitting.
Second, compared with existing NMF models such as minimum-volume NMF \cite{ang2018volume,leplat2020blind}, which were shown to exhibit rank-finding ability \cite{leplat2019minimum}, SON-NMF is applicable to rank-deficient matrices.
\end{itemize}

\item \textbf{Irreducible computational complexity.}
As computing $\rk_+$ is NP-hard, the SON approach, as a ``work-around'' method to estimate $\rk_+$, cannot reduce computational complexity.
We prove (Theorem~\ref{thm:irreducible}) that the complexity of the SON term is \textit{almost irreducible}.
Precisely, we show that in the best case, 
to recover the $r^*$ columns of the true $\W^*$ using $\W$ obtained from SON-NMF with rank $r>r^*$, the complexity of the SON term cannot be reduced from $r(r-1)/2$ to below $r(r-\lceil r/r \rceil)/2$.

\item \textbf{Fast algorithm by proximal-average.}
Solving \eqref{eqn:sonnmf} is nontrivial: the $\W$-subproblem is nonsmooth, non-separable, and non-proximable, so existing proximal-based methods \cite{tseng2009coordinate,xu2013block,razaviyayn2013unified,bolte2014proximal,le2020inertial} cannot efficiently solve the problem.
For non-proximal problems, dual approaches such as Lagrange multipliers or ADMM are typically used. 
However, SON-NMF involves $\cO(r^2)$ non-proximal terms, and this complexity is irreducible (Theorem~\ref{thm:irreducible}). 
Therefore, dual and 2d-order methods are inefficient due to their high per-iteration cost.
We propose a low-cost proximal average approach \cite{yu2013better} based on the Moreau-Yosida envelope \cite{bauschke2008proximal}. 
\end{itemize}
We review the literature, focusing on the background and motivation of this work.
\paragraph{Review of NMF: minimum-volume and rank-deficiency}
SON-NMF is related to minimum-volume (minvol) NMF \cite{ang2018volume,ang2019algorithms}.
Recently, it was observed in \cite{leplat2019minimum} that when using volume regularization in the form of $\log\det(\W^\top \W + \delta \I_r)$, minvol NMF applied to a rank-deficient matrix $\M$ (i.e., when the $r$ parameter is overestimated) can zero out the extra components in $\W$ and $\H$.
This phenomenon was also observed in audio blind source separation \cite{leplat2020blind}, where a rank-7 factorization was applied to a dataset with 3 sources: the minvol NMF was able to set the redundant components to zero.
However, minvol NMF is not suitable for rank-deficient $\W$: if $\delta = 0$, then $\log\det(\W^\top \W) = \log 0 = -\infty$.
Even if $\delta \neq 0$, a rank-deficient $\W$ provides little information in the log-det term.
Furthermore, in \cite{leplat2020blind}, when using an overestimated rank in minvol NMF, it is the redundant components in $\H$ that are set to zero, rather than those in $\W$.
We remark that this rank-revealing property of minvol NMF motivated the first author to propose SON-NMF.

\paragraph{Review of SON}
SON was originally proposed in \cite{pelckmans2005convex,lindsten2011clustering} for clustering.
Because minimizing SON$(\W)$ forces the pairwise differences $\w_i - \w_j$ to be small, SON is also referred to as a ``fusion penalty'' \cite{hocking2011clusterpath}.
Later, \cite{niu2016nonsmooth} considered SON with $0 < p < 1$, and more recently, \cite{jiang2020certifying} showed that SON-based clustering can provably recover Gaussian mixtures under certain assumptions.
SON$_{2,0}$ has also been applied in graph trend filtering \cite{huang2025inhomogeneous}.
We note that these works differ from SON-NMF: they involve single-variable problems, whereas NMF is a bi-variate, nonconvex problem with nonnegativity constraints.

\paragraph{SON solution approaches}
The approach we propose to solve the SON problem differs from existing methods such as quadratic programming with convex hull \cite{pelckmans2005convex}, active-set methods \cite{hocking2011clusterpath}, interior-point methods \cite{lindsten2011clustering}, trust-region methods with smoothing \cite{niu2016nonsmooth}, Lagrange multiplier methods \cite[12.3.8]{beck2017first}, and semi-smooth Newton methods \cite{yuan2018efficient}.
These approaches were all designed for single-variable clustering problems (i.e., involving $\W$ only) without nonnegativity constraints.
In contrast, we leverage the proximal average \cite{bauschke2008proximal,yu2013better}, which is computationally inexpensive to compute, with a per-iteration cost of $\cO(m)$, where $m$ is the dimension of $\w_j$.
This significantly lowers the per-iteration cost for SON in our setting.
All the aforementioned methods are either unable to handle the SON problem with nonnegativity constraints on $\W$ or incur higher per-iteration costs.
See details in \cref{sec:algo21W}.

\paragraph{History: the geometric median and the Fermat-Torricelli-Weber problem}
SON was proposed in the 2000s \cite{pelckmans2005convex, hocking2011clusterpath, lindsten2011clustering}, it is closely related to an older problem known as the Fermat-Torricelli-Weber problem \cite{krarup1997torricelli,nam2014nonsmooth}, also called as the geometric median \cite[E.g.3.66]{beck2017first}.
The analysis of the geometric median does not directly apply to SON-NMF, but it provides a geometric interpretation: SON-NMF produces an $r^*$-cluster of points that minimizes the geometric median distance to the dataset.

\paragraph{Rank estimation in NMF}
Existing methods for rank estimation in NMF are not applicable in the setting of this paper.
Algebraic approaches, such as fooling sets \cite{cohen1993nonnegative} and the $f$-vector \cite{dewez2021geometric}, only provide loose bounds on $\rk_+(\M)$ and are computationally expensive to implement.
Statistical approaches \cite{tan2012automatic, squires2017rank, esposito2020nmf} assume that $\W$ and $\H$ follow predefined distributions or require heavy post-processing.
SON-NMF makes none of these assumptions and requires no post-processing.

\paragraph{A ``drawback'' of SON-NMF}
Finding $\rk_+$ in NMF is NP-hard, and the search space of $r$ in NMF is the set of natural numbers $\IN$, which is countably infinite.
In SON-NMF, we do not need to estimate the rank $r$, but we must provide a regularization parameter $\lambda$, whose search space is the set of nonnegative real numbers $\IR_+$.
By Cantor's diagonal argument \cite{cantor1890ueber}, the cardinality of the real numbers is uncountably infinite.
Hence, theoretically, SON-NMF replaces the search space $\IN$ of NMF with the much larger space $\IR_+$, suggesting that SON-NMF could be even more difficult to solve than the already NP-hard NMF.
We remark, however, that this is not an issue in practice: many datasets are hierarchically clustered in the latent space, so a simple tuning of $\lambda$ is sufficient for SON-NMF to recover the true rank.

\paragraph{Paper organization}
We present the theory of SON-NMF \cref{sec:theo}.
We describe how to solve SON-NMF in \cref{sec:algo21} and \cref{sec:algo21W}.
In \cref{sec:exp}, we show experimental results, and \cref{sec:conclusion} concludes the paper.

\paragraph{Notation}
The notation ``$\{x,y\}$ denotes $\{X,Y\}$'' means that $X$ is denoted by $x$ and $Y$ by $y$, respectively (resp.).
We use $\{\IR, \IR_+, \overline{\IR},\IRm, \IRmn\}$ to denote $\{$reals, nonnegative reals, extended reals, $m$-dimensional reals, $m$-by-$n$ reals$\}$,
we use $\{$lowercase italic, bold lowercase italic, bold uppercase letters$\}$ to denote $\{$scalar, vector, matrix$\}$.
Given a matrix $\M$, we denote $\{\m^i, \m_j\}$ the $\{i$th row, $j$th column$\}$ of $\M$.
Given a convex set $C \subset \IRn$, the indicator function of $C$ at $\x$ is defined as $\iota_C(\x) = 0$ if $\x \in C$ and $\iota_C = +\infty$ if $\x \notin C$, 
and $\proj_{C}(\x)$ denotes the projection of $\x$ onto $C$.
The projection of $\{\v \in \IRn, \V \in \IRmn\}$ onto the nonnegative orthant $\{\IRn_+, \IRmn_+\}$ is denoted by the element-wise max operator $\{ [\v]_+, [\V]_+\}$.
Lastly, $\Delta^r \subset \IR^r$ denotes the unit simplex, and $\ones_r \in \IR^r$ is the vector of ones.

\begin{remark}
The constraint on $\H$ removes the scaling ambiguity of the factorization.
I.e., for $\M = \W_1 \H_1$, there does not exist a diagonal matrix $\D$ such that $\M = (\W_1 \D)(\D^{-1} \H_1) \eqqcolon \W_2 \H_2$ with $\W_1 \neq \W_2$ and $\H_1 \neq \H_2$.
\end{remark}

\section{Theory of SON-NMF}\label{sec:theo}
In this section, we present the theory of SON$_{2,1}$-NMF.
We first examine SON$_{2,1}$ in detail, then motivate why reducing its computational complexity is desirable, and provide a bound showing that this complexity is essentially irreducible.

\subsection{SON$_{2,1}$ has $r^2$ terms and its minimum occurs at maximal cluster imbalance}
Let $\X = [\x_1, \x_2, \dots]$ have five columns.
The pairwise differences $\x_i - \x_j$ in SON yield $5^2 - 5 = 20$ pairs.
In general, if $\X$ has $r$ columns, there are $r(r-1)$ ordered pairs $(\x_i, \x_j)$.
By symmetry, $|\x_i - \x_j|_2 = |\x_j - \x_i|_2$, so we consider only the $r(r-1)/2$ \textit{distinct} pairs in SON.
We can describe this in graph-theoretic terms.
Let $G(V,E)$ be a simple, undirected, unweighted graph with $|V|$ nodes and $|E|$ edges, and let $K_r$ be the complete graph on $r$ nodes.
Then 
\[
\text{SON}_{2,1}(\X) = \sum_{(i,j) \in E(K_r)} \| \x_i - \x_j \|_2
\]
Since $|E(K_r)| = r(r-1)/2$, SON$_{2,1}$ contains $\cO(r^2)$ terms.

Returning to the example with five columns, let $\X$ be a rank-3 matrix with three clusters having centers $\c_1, \c_2, \c_3$:
$ \X
= [\x_1 ~|~ \x_2 ~|~ \x_3 ~ \x_4 ~ \x_5]
= [ \c_1  ~|~ \c_2 ~ |~ \c_3 ~ \c_3 ~ \c_3 ]
$.
Then $\text{SON}_{2,1}(\X) = \|\c_1 - \c_2\|_2 + 3 \|\c_1 - \c_3\|_2 + 3 \|\c_2 - \c_3\|_2$.
The graph $K_5$ and the corresponding pairwise differences are illustrated below.

\begin{minipage}{0.3\textwidth}
\centering
\begin{tikzpicture}
\graph { subgraph K_n [n=5,clockwise,radius=1.15cm] };
\end{tikzpicture}    
\end{minipage}
\begin{minipage}{0.57\textwidth}
\begin{tabular}{c|c|c|ccc}
  & 1 & 2 & 3 & 4 & 5 \\ \hline
1 & 0 & $\c_1 -\c_2$ & $\c_1-\c_3$ & $\c_1 -\c_3$ & $\c_1 - \c_3$ 
\\ \hline
2 &   & 0            & $\c_2 -\c_3$            & $\c_2 -\c_3$ & $\c_2 - \c_3$ 
\\ \hline
3 &   &   & 0        & 0& 0 
\\
4 &   &   &   & 0  & 0
\\
5 &   &   &   &    & 0
\end{tabular}    
\end{minipage}

Now we generalize.
Let $\X$ have $r$ columns and $r^* \leq r$ clusters $C_1, \dots, C_{r^*}$ with centers $\c_1, \dots, \c_{r^*}$.
Let $|C_i|$ denote the size of cluster $C_i$.
Since $|C_1| + \cdots + |C_{r^*}| = r$, we have
\[
\begin{array}{rcl}\displaystyle
\hspace{-1mm}
\text{SON}_{2,1}(\X) = \hspace{-1mm}
\sum_{(i,j) \in K_r} \hspace{-2mm} |C_i| |C_j|  \big\| \c_i - \c_j \big\|_2
&\leq&\displaystyle\hspace{-2mm}
\Big(\max_{i \in [r]} |C_i|\Big)
\Big(\max_{i,j} \big\| \c_i - \c_j \big\|_2 \Big)  
\hspace{-2mm}
\sum_{(i,j) \in K_r}  |C_j| 
\\
\hspace{-1mm}&\leq&\hspace{-2mm}\displaystyle
\Big(\max_{i \in [r]} |C_i|\Big)
\Big(\max_{i,j} \big\| \c_i - \c_j \big\|_2 \Big) r,
\end{array}
\]
giving a stopping criterion for the algorithm: we have $r$ as input, so we just need to track the product $\displaystyle \big(\max_{i \in [r]} |C_i|\big)
\big(\max_{i,j} \big\| \c_i - \c_j \big\|_2 \big)$ for convergence.  
Furthermore, from the inequality, we can focus on the cluster sizes rather than the norms $\|\c_i-\c_j\|_2$, leading to the following lemma that characterizes the theoretical minimum of SON$_{2,0}$ as a proxy for SON$_{2,1}$:
\begin{lemma}[Maximal cluster imbalance gives the minimum of SON$_{2,0}$]\label{lem:SON20min}
For an $n$-column matrix $\X$ with $K$ clusters $C_1,\dots,C_K$, where all $\x_i \in C_i$ take the centroid $\c_i$,
\[
\textrm{SON}_{2,0}(\X) = \sum_{i,j} |C_i| |C_j| 
\]
achieves its minimum when one cluster contains $n-K+1$ columns of $\X$, and the remaining $K-1$ clusters each have size one.
\begin{proof}
Directly by $|C_1| + \dots + |C_K| = n$ and basic inequality manipulations.
\end{proof}
\end{lemma}

\paragraph{Remarks of Lemma~\ref{lem:SON20min}}
\begin{enumerate}[leftmargin=12pt]
\item Since the $\ell_1$-norm is a tight convex relaxation of the $\ell_0$-norm (over the unit ball), Lemma~\ref{lem:SON20min} for the SON$_{2,0}$ term provides a theoretical minimum for the SON$_{2,1}$ term when the input matrix lies within a unit ball.

\item For any cluster $C_i$, the smallest cluster size is $1$, so the SON$_{2,0}$ term (and similarly the SON$_{2,1}$ term) cannot ``miss'' a weak component in the data if one exists.
This behavior is observed in experiments (see Figs.~\ref{fig:urban_result},~\ref{fig:exp2}).
Furthermore, if the cluster centers $\c_i$ are approximately equidistant ($\|\c_i - \c_j\|_2 \approx \|\c_j - \c_k\|_2$), maximal cluster imbalance naturally occurs in applications (see Figs.~\ref{fig:swimmer}, \ref{fig:region}, \ref{fig:exp3}, \ref{fig:urban_result}).
\end{enumerate}

\subsection{SON complexity is irreducible}
We now see that SON has $\cO(r^2)$ terms.
In practice, SON-NMF often uses a large input rank $r$-potentially as large as the data dimensions $m$ or $n$-to estimate the true rank $r^*$.
This makes the SON term computationally expensive.
A natural question arises: can we reduce the complexity of the SON term by removing some edges in $K_r$, thereby lowering the per-iteration cost of SON-NMF while preserving recovery performance?
The answer is negative, as stated in Theorem~\ref{thm:irreducible}: the complexity of the SON term is essentially irreducible.

\begin{remark}
There are prior works explored similar ideas.
For example, \cite[page 2]{yuan2018efficient} mentions approaches using $k$-nearest neighbors.
However, these are data-dependent methods that leverage the data to learn a graph structure for reducing the complexity of the SON term.
Our focus is different. 
We investigate the possibility of reducing the complexity of the SON term purely from a graph-theoretic perspective, independent of the data. 
That is, we are interested in whether a sparsest subgraph exists such that the reduced SON is ``functionally the same'' as the full-SON. Theorem~\ref{thm:irreducible} shows that such sparsest subgraph basically does not exist.
\end{remark}

We introduce notation.
Let $r^*$ be the true NMF rank of $\W$.
For a graph $G(V,E)$, let $u,v \in V$ be two nodes connected by an edge, i.e., $(u,v) \in E$.
The notation $G \setminus (u,v)$ denotes the subgraph obtained by removing the edge $(u,v)$ from $G$.
A \emph{graph partition} of $G$ is a set of subgraphs $S_1, S_2, \dots$ such that the vertex sets $V(S_i)$ form a mutually exclusive partition of $V(G)$.

We now state a trivial fact.

\begin{lemma}\label{lem:partition}
Let $\W$ have NMF rank $r^*$.
Then the graph generated by the columns of $\W$ satisfies the following property: for every partition of its nodes into $r^*$ subgraphs, each subgraph must be connected.
\end{lemma}
This lemma can be easily proved by contradiction.
Next we have a lemma.
\begin{lemma}
The only graph satisfying the condition of Lemma~\ref{lem:partition} is the complete graph.
\begin{proof}
Let $G$ be a graph whose nodes correspond to the columns of $\W$.
Suppose some edge $(u,v)$ is omitted from $G$.
Then it is possible for the `true' partition to place $u$ and $v$ in the same subgraph, while the remaining nodes of $G \setminus (u,v)$ are divided arbitrarily among the other $r^*-1$ subgraphs.
In this case, $u$ and $v$ are no longer directly connected, violating the connectivity requirement.
Since $u$ and $v$ were arbitrary, no edge can be omitted from $G$, and thus $G$ must be complete.
\end{proof}
\end{lemma}
This lemma implies that, apart from the complete graph, no other graph structure satisfies the connectivity condition.
The following theorem quantifies how many edges can be removed from the complete graph, which tells essentially ``none''.

\begin{theorem}\label{thm:irreducible}
Let $(\W^*, \H^*) = \NMF(\M,r^*)$ and let $(\W, \H) = \text{SON-NMF}(\M,r)$ with $r \ge r^*$.
If we aim to recover $\W^*$ using $r^*$ clusters among the $r$ columns of $\W$, the number of pairwise difference terms $|\w_i - \w_j|_2$ in SON cannot be reduced below
\[
\dfrac{r}{2}\left(r - \left\lceil \dfrac{r}{r^*} \right\rceil\right).
\]
\begin{proof}
Represent the $r$ columns of $\W$ as nodes of a simple undirected graph $G(V,E)$, where each edge $(u,v)$ corresponds to the term $|\w_i - \w_j|_2$.
Recovering $\W^*$ by $\W$ with fewer SON terms  can be translated as 
\begin{equation}\label{st:reduction}
\begin{array}{l}
\text{we can identify $r^*$ disjoint clusters in $G$ with $|V|=r$}  
\\
\text{using a subgraph of $K_r$ with fewer edges.}  
\end{array}
\end{equation}
We are going to show that the statement \eqref{st:reduction} is true, and at best such an improvement is from $r(r-1)/2$ to $r(r-\lceil r/r^* \rceil)/2$.

Assuming, in the best case that, each of these $r^*$ columns of $\W^*$ is corresponds to exactly $\lceil r / r^* \rceil$ nodes in $\W$, represented by the nodes in the graph $G$.
For any node $v \in V$, let $S(v) \subset V$ be the set of nodes disconnected\footnote{I.e., there is no path between $u \in S(v)$ and $v$} from $v$, and let $T$ be a nonempty subset of $S(v)$.
Then the recovery of the $r^*$ clusters in $G$ is impossible if a cluster in $G$ is of the form $\{v\} \cup T$.
The negation of this very last statement gives:
\[
\begin{array}{l}
\text{
To recover the $r^*$ clusters for all subset of nodes of size at least $r/r^*$,
}
\\
\text{
we need $ |T| < r/r^*$ for any such $T$.
}
\end{array}
\]
The inequality $ |T| < r/r^*$ holds for all subset $T$ of $S(v)$, this implies $ |S(v)| < r/r^*$.
I.e., $v$ has to connect to at least  $ r -  \lceil  r/r^* \rceil $
other nodes $u \notin S(v) $ in the graph $G$.
This connectivity holds for every node $v \in V$, meaning that the minimum number of edges required is $r\big( r - \lceil  r/r^* \rceil \big)/2$.
\end{proof}
\end{theorem}
Theorem~\ref{thm:irreducible} shows that the number of edges in SON cannot be significantly reduced from $r(r-1)/2$.
We define the reduction factor $R(r^*,r)$ as
\[
\begin{array}{rcl}
R(r;r^*)
&\coloneqq&
\dfrac{\text{full number of terms} - \text{reduced number of terms}}{\text{full number of terms}}
\vspace{3mm}
\\
&=& \displaystyle
\frac{\frac{r}{2}( r - 1 )
-
\frac{r}{2}\big( r - \big\lceil r/r^* \big\rceil\big)}
{\frac{r}{2}( r - 1 )}.
\end{array}
\]
The following lemma tells that the reduction factor is small.
\begin{lemma}
For fixed $r^*$, the value $R(r^*,r)$ approaches to $1/r^*$ as $r \to \infty$.
\begin{proof}
Take the limit of $R(r;r^*)$ gives
$\displaystyle
\lim_{r \rightarrow \infty} R(r;r^*) 
=
\lim_{r \rightarrow \infty}
\frac{\big\lceil r/r^* \big\rceil  -1}{r-1}
=
\lim_{r \rightarrow \infty}
\frac{\big\lceil r/r^* \big\rceil}{r-1}
$.
Using $r \leq \lceil r \rceil \leq r+1 $, we have
$\displaystyle
\lim_{r \rightarrow \infty} \dfrac{ r/\lceil r^* \rceil }{r-1}
\leq
\lim_{r \rightarrow \infty} R(r;r^*) 
\leq
\lim_{r \rightarrow \infty} \dfrac{ (r+1)/\lceil r^* \rceil}{r-1}
$.
By squeeze theorem, 
$\displaystyle \lim_{r \rightarrow \infty} R(r;r^*)  = 1/ r^*
$.
\end{proof}
\end{lemma}
The lemma shows that the SON term's complexity can only be reduced marginally.
For $r^* \ge 3$ (NMF is trivial for $r^* \le 2$ \cite{gillis2020nonnegative}), the maximum reduction is about 33\%.
This reduction quickly decreases as $r$ or $r^*$ increases. 
For example, with $(r,r^*)=(1000,25)$, using 1000 nodes to find 25 clusters, the number of edges in $K_{1000}$ can be reduced by at most 5\%, from $|K_{1000}| = 499{,}500$ to $500(1000-\lceil 1000/25\rceil)=480{,}000$.

\section{BCD algorithm and the H-subproblem}\label{sec:algo21}
We now discuss solving the nonsmooth, nonconvex, nonseparable, and non-proximable problem \eqref{eqn:sonnmf} via block coordinate descent (BCD) \cite{hildreth1957quadratic, wright2015coordinate}.
Let $k$ denotes the iteration counter.
Starting with an initial guess $(\W_1,\H_1)$, we perform alternate update :
$\H_{k+1} \leftarrow \textrm{update}(\H_k;\W_k)$, 
$\W_{k+1} \leftarrow \textrm{update}(\W_k;\H_{k+1})$, where \textrm{update}$()$ denotes an approximate solution to the respective subproblem.
In this section, we describe the BCD framework and the update for $\H$.
The $\W$-subproblem is discussed in the next section.

Algorithm~\ref{algo:BCDframework} shows the pseudo-code of the BCD method for solving SON-NMF.

\begin{algorithm}[!h]\label{algo:BCDframework}
\DontPrintSemicolon
\KwInput{$\M,\W_1, \H_1,\lambda, \gamma $} 
\For{$k=1,2,\dots$}{
$
\H_{k+1} = 
\proj_{\Delta^r}\Big( \Q \H_k + \R \Big)
$
with 
 $\Q  = \I_n - \W_k^\top \W_k / \| \W_k^\top \W_k\|_2$  and $\R = \W_k^\top \M / \| \W_k^\top \W_k \|_2$
\;
\For{$\ell = 1,2,\dots,\ell_{\text{max}}$, (e.g., $10$)}{
$\W_{k+1} = \textrm{update}(\W_k; \H_{k+1}, \M, \lambda, \gamma )$, see \cref{sec:algo21W}.
}
}
\caption{(Inexact) BCD for solving SON-NMF}
\end{algorithm}

We now explain Step 2 in Algorithm~\ref{algo:BCDframework}.

\paragraph{H-subproblem: projection onto unit simplex}\label{sec:BCD:subsec:H}
The step $\textrm{update}(\H_k;\W_k)$ solves the subproblem on $\H$, which consists of $n$ parallel problems:
\begin{equation}\label{prob:subH}
\argmin_{\h_1,\dots,\h_n} 
\dfrac{1}{2} \sum_{j=1}^n \| \W_k \h_j - \m_j \|_2^2 
~~\st~~ \h_j \in \Delta^r  ~\text{for }j=1,2,\dots,n,
\end{equation}
where $\Delta^r \coloneqq \big\{ \x \in \IR^r_+ : \sum_i x_i \leq 1 \big\}$.
Each column $\h_j$ solves a a constrained least-squares:
\begin{equation}\label{prob:agmin_h_sub}
\argmin_{\x \in \Delta^r} 
f(\x)~=~
\dfrac{1}{2} \| \A \x - \b \|_2^2
~=~ \dfrac{1}{2}\langle \A^\top\A\x, \x \rangle - \langle \A^\top \b, \x \rangle
,
\end{equation}
where $\x$ is the variable $\h_j$, and $\A = \W^\top\W$ with $\b = \W^\top \m_j$.
We use proximal gradient method (details in the next section) to update $\x$ in \eqref{prob:agmin_h_sub} iteratively as 
\begin{equation}\label{updt:agmin_h_sub}
\x_k^{\ell+1}
=
\proj_{\Delta^r}\bigg(
\x_k^\ell - \dfrac{ \A^\top \A \x_k^\ell - \A^\top \b}{\| \A^\top \A \|_2}
\bigg),
\end{equation}
where $\x_k^\ell$ is the variable at iteration-$k$ and inner-iteration-$\ell$. 
For low per-iteration cost, we typically set $\ell= 1$.

\paragraph{Projection}
$\proj_{\Delta^r}( \x )$ projects $\x \in \IR^r$ onto the unit simplex $\Delta^r$ with cost $\cO(r \log r)$ \cite{condat2016fast}, due to sorting when computing the Lagrange multiplier.

\paragraph{Matrix update}
The column-wise updates can be combined into a matrix update:
\[
\H_{k+1}
= 
\proj_{\Delta^r}\Bigg(
\H_k - 
\dfrac{ 
\W_k^\top \W_k \H_k - \W_k^\top \M
}
{
\| \W_k^\top \W_k \|_2
}
\Bigg),
\]
with $\proj_{\Delta^r}$ applied in parallel to each column.
The total cost is $\cO(nr \log r)$.
For $r \approx n$, the cost is $\cO(n^2 \log n)$, this partly explains why 2nd-order methods are impractical.
Below we give another reason for not considering 2nd-order method for updating $\H$: the $\W$ is multicollinear.

\paragraph{Impact of $\W$-multicollinearity}
As SON encourages multicollinearity in $\W$, so $\W$ and $\W^\top \W$ may be ill-condotioned.
This affects Problem\,\eqref{prob:subH}:
\begin{enumerate}[leftmargin=12pt]
    \item The problem may not be strongly convex, possibly yielding multiple global minima.

    \item Nesterov acceleration \cite{nesterov2003introductory} becomes less effective due to the large condition number.  

    \item 2nd-order method are infeasible because $(\W^\top_k\W_k)^{-1}$ may not exists.

    \item Duality-based tools (e.g., for stopping criteria) cannot be applied efficiently.
\end{enumerate}

\section{Proximal averaging on the W-subproblem}\label{sec:algo21W}
We now focus on solving the $W$-subproblem, the line $\textrm{update}(\W; \H, \M, \lambda, \gamma )$ in Algorithm~\ref{algo:BCDframework}:
\begin{equation}\label{eqn:subproblemW}
\argmin_{\W} \, F(\W) 
~\coloneqq~ 
\dfrac{1}{2}\| \W\H - \M \|_F^2 + \lambda \sum_{i \neq j} \| \w_i - \w_j \|_2
+ \gamma \sum_{j=1}^r \big\| \max\{ -\w_j , \zeros\} \big\|_1.
\end{equation}
The function $F(\W)$ in \eqref{eqn:subproblemW} has the following properties:
\begin{itemize}[leftmargin=12pt]
\item Convex and continuous: All terms are norms under convex-preserving maps.
$\| \w_i - \w_j \|_2$, $\| \max\{ -\w_j , \zeros\}\|_1$ are $1$-Lipschitz.
\item Nonsmooth: $\| \w_i - \w_j \|_2$ is non-differentiable at $\w_i = \w_j$ and $\| \max\{ -\w_j , \zeros\}\|_1$ is non-differentiable at negative entries.
\item Non-separable: $\w_i,\w_j$ are lumped together in SON, so $F(\W)$ cannot be split into independent column-wise functions.
\item Non-proximable: The prox operator for $ \lambda \sum \| \w_i - \w_j \|_2 + \gamma \sum \| \max\{ -\w_j , \zeros\}\|_1$ has no closed-form solution and cannot be efficiently computed.

\item Not dual-friendly: Introducing dual variables (e.g., for ADMM) increases the number of variables from $r$ to $r^2$, which is impractical for large $r$. 

\item Not 2nd-order friendly:  Computing Hessians or Newton steps is prohibitive, with per-iteration cost $\cO(m^4)$ to $\cO(m^5)$ if $r \sim m$.
\end{itemize}
Because of these properties, standard proximal gradient methods \cite{tseng2009coordinate,xu2013block,razaviyayn2013unified,bolte2014proximal,le2020inertial} are inefficient.
Instead, we solve \eqref{eqn:subproblemW} using the Moreau-Yosida envelope with proximal averaging \cite{yu2013better}, which avoids parameter tuning required in inexact proximal \cite{schmidt2011convergence} or smoothing methods \cite{nesterov2005smooth}.

\begin{remark}
$\sum_{i} \| \max\{ -\w_i , \zeros\}\|_1$ enforces $\W \geq \zeros$ if $\gamma > 0$ is sufficiently large. 
\end{remark}

\paragraph{Column-wise update}
We solve \eqref{eqn:subproblemW} column by column.
Consider the $j$th rank-1 component $\w_j\h^j$.
Let $\M_j = \M - \W_{-j}\H^{-j}$ where $\W_{-j}$ is $\W$ without column $\w_j$ and $\H^{-j}$ is $\H$ without row $\h^j$.
The subproblem \eqref{eqn:subproblemW} on $\w_j$ becomes
\begin{equation}\label{eqn:subproblemwj}
\w_j^* 
~\coloneqq~ 
\argmin_{\w}  \dfrac{\| \h^j \|_2^2}{2} \| \w \|^2_2 - \langle \M_j \h^j{^\top}, \w \rangle + \lambda \sum_{i \neq j} \| \w - \w_i \|_2 
+ \gamma \| \max\{ -\w, \zeros\}\|_1.
\end{equation}
which can be cast in the general form
\begin{equation}\label{min:phipsi}
\argmin_{\x}\, \phi(\x) + \psi(\x),~~~\textrm{where }~
\psi(\x) \coloneqq \sum_{i=1}^N \alpha_i \psi_i(\x),
\end{equation}
where $\phi : \IRm \rightarrow \IR$ is closed, proper, convex, and smooth.
Each $\psi_i : \IRm \rightarrow \barIR$ is closed, proper, convex, and possibly nonsmooth. Coefficients $\alpha_i \geq 0$ are normalized ($\sum \alpha_i = 1$) by $\lambda$ and $\gamma$.
Note that $\psi_i$ are non-separable, i.e., they share the same global variable $\x$.

\paragraph{Proximal gradient method}
A standard approach to solve minimization \eqref{min:phipsi} is the proximal gradient method
\cite{passty1979ergodic,fukushima1981generalized,combettes2005signal}, in update under a  stepsize $\mu > 0$ as
$\x^+ = \rmP_{\psi}^{\mu} \big( \x - \mu \nabla \phi(\x)  \big)$,
where $\rmP_{\psi}^{\mu}$ is the proximal operator of $\psi$ (see \eqref{def:PM}).
By $\psi(\x) \coloneqq \sum_{i=1}^N \alpha_i \psi_i(\x)$ in \eqref{min:phipsi}, we have $\rmP_{\psi}^{\mu} = \rmP_{\sum \alpha_i \psi_i}^{\mu}$ in which intractable, this is what we mean by $\psi$ being ``non-proximable''.
To address this, we employ the proximal average 
\cite{bauschke2008proximal,yu2013better}.
Below we first give the background of proximal average for solving \eqref{min:phipsi}, then we explain its application to \eqref{eqn:subproblemwj}.

\subsection{Proximal average}
Given a point $\v \in \IRn$, a convex, closed, proper function $f: \IRn \rightarrow \IR \cup \{+\infty\}$, and a parameter $\mu > 0$, the proximal operator of $f$ at $\v$, denoted as $\rmP_{f}^{\mu}(\v)$, 
and the Moreau-Yosida envelope of $f$ at $\v$, denoted as $\rmM_{f}^{\mu}(\v)$, are defined as \\
\begin{minipage}{.567\linewidth}
\begin{equation}\label{def:PM}
\begin{array}{l}
\rmP_{f}^{\mu}(\v) 
\coloneqq \displaystyle \argmin_{\bxi} f(\bxi) + \frac{1}{2\mu} \| \bxi - \v \|_2^2,
\\
\rmM_{f}^{\mu}(\v) 
\coloneqq \displaystyle \min_{\bxi} f(\bxi) + \frac{1}{2\mu} \| \bxi - \v \|_2^2.
\end{array}
\end{equation}
\end{minipage}
\hfill
\begin{minipage}{.5\linewidth}
 \setlength{\interspacetitleruled}{0pt}
 \setlength{\algotitleheightrule}{0pt}
\begin{algorithm}[H]
\DontPrintSemicolon
\For{$k=1,2,\dots$}{
$\bar{\x} = \x_{k}  - \mu \nabla \phi(\x_{k} )$ \;
$\displaystyle \x_{k+1} = \sum_{i=1}^N \alpha_i \rmP_{\psi_i}^{\mu} (\bar{\x})$
}
\caption{Proximal average}
\label{algo21:prox_avg_x}
\end{algorithm}
\end{minipage}
\\
The idea of proximal average is that computing $\rmP_{\psi}^{\mu} = \rmP_{\sum \alpha_i \psi_i}^{\mu}$ directly is hard, while the individual $\rmP_{\psi_i}^{\mu}$ is easy to compute.
We therefore approximate $\rmP_{\psi}^{\mu}$ by $\sum_{i=1}^N \alpha_i \rmP_{\psi_i}^{\mu}$.
Algorithm~\ref{algo21:prox_avg_x} implements this approach to solve \eqref{min:phipsi}.
Under the assumptions that $\phi$ is $L$-smooth and $\psi_i$ are all $M_i$-Lipschitz, the sequence $\{\x_k\}_{k \in \IN}$ produced by Algorithm~\ref{algo21:prox_avg_x} converges to the minimizer of 
\[
\argmin_{\x} \phi(\x) + A(\x), ~\textrm{where }~
\rmM_A^\mu = \sum_i \alpha_i \rmM_{\psi_i}^\mu,
\tag{$\#$}
\]
with $A$ called the proximal average of $\{\psi_1,\dots,\psi_n\}$ \cite{yu2013better}.
Moreover, we have $0 \leq \psi - A \leq \frac{\mu}{2} \sum_i \alpha_i M_i^2 < +\infty$ which implies that an $\epsilon-$solution for ($\#$) is an $2\epsilon$-solution for \eqref{min:phipsi}.

\subsection{Update on w}
We now explain how to use proximal average (Algorithm~\ref{algo21:prox_avg_x}) to solve the W-subproblem.
First, let $\sigma = (r-1)\lambda + \gamma$ be a normalization factor and rewrite \eqref{eqn:subproblemwj} as 
\begin{equation}\label{eqn:subproblemwj_2}
\argmin_{\w} 
\dfrac{\| \h^j \|_2^2}{2} \| \w \|^2_2 
- \langle \M_j \h^j{^\top}, \w \rangle 
+ \sigma 
\Big( \sum_{1 \leq i \neq j \leq r}^{r} \frac{\lambda}{\sigma } \| \w - \w_i \|_2 
+ \dfrac{\gamma}{\sigma } \| \max\{ -\w, \zeros\}\|_1
\Big).
\end{equation}
Since $\argmin F = \argmin \alpha F$ for all $\alpha >0$, we scale by $1/\sigma$ to get 
\begin{equation}\label{eqn:subproblemwj_3}
\argmin_{\w} 
\underbrace{\dfrac{\| \h^j \|_2^2}{2\sigma} \| \w \|^2_2 
- \Big\langle \dfrac{\M_j \h^j{^\top}}{\sigma}, \w \Big\rangle }_{\phi}
+ \sum_{1 \leq i \neq j \leq r}^{r} \frac{\lambda}{\sigma } \| \w - \w_i \|_2 
+ \dfrac{\gamma}{\sigma } \| \max\{ -\w, \zeros\}\|_1,
\end{equation}
which satisfies the assumptions required for proximal averaging.

The gradient of $\phi$ is $
\nabla \phi(\w) = \| \h^j\|_2^2\w / \sigma - \M_j \h^j{^\top}/\sigma$
and it is $(\|\h^j\|_2^2/\sigma)$-Lipschitz.
Thus, the gradient step in Algorithm~\ref{algo21:prox_avg_x} becomes
\[
\overline{\w} 
~=~
\w - \dfrac{1}{L} \nabla \phi(\w)
~=~
\w - \dfrac{1}{\| \h^j\|_2^2 /\sigma} \Big(
\dfrac{\| \h^j\|_2^2}{\sigma} \w
-  
\dfrac{\M_j \h^j{^\top}}{\sigma}
\Big)
~=~
\dfrac{\M_j \h^j{^\top}}{\| \h^j\|_2^2}.
\]

Next we recall three useful lemmas for computing the prox of each nondifferentiable terms:
\begin{lemma}[Scaling]
If $\nu > 0, \mu > 0$ then $ \rmP^{\mu}_{\nu \psi} = \rmP^{\nu \mu}_\psi$.
\end{lemma}

\begin{lemma}
The proximal operator of $\| \x - \c \|_2$ with parameter $\mu$ is 
\[
\rmP^{\mu}_{\| \x - \c \|_2}(\v) 
= 
\v - \dfrac{\v-\c}{\max\left\{\, 1,  \big\|  \frac{\v-\c}{\mu}\big\|_2  \right\}}.
\]
\end{lemma}

\begin{lemma}
Let $\ones$ be the vector of ones, the proximal operator of $\mu \| \max\{-\x , \zeros \} \|_1$ has the closed-form expression  
$\textrm{median}\big( \v + \mu\ones , \zeros, \v \big)$, i.e.,
\[
\Big[\rmP^{1}_{\mu \| \max\{-\cdot , \zeros \} \|_1}(\v) \Big]_i 
~=~
\begin{cases}
v_i + \mu & v_i + \mu < 0,
\\
0 & v_i \leq 0 \leq v_i + \mu,
\\
v_i & v_i > 0.
\end{cases}
\]
\end{lemma}
Based on the three lemmas, the proximal step for the SON terms is
\[
\rmP^{\frac{1}{L_j}\frac{\lambda}{\sigma}}_{\| \cdot - \w_i \|_2}(\bar{\w})
~=~
\rmP^{\frac{\lambda}{\| \h^j\|_2^2}}_{\| \cdot - \w_i \|_2}(\bar{\w})
~=~
\bar{\w} - \dfrac{\bar{\w} - \w_i}{\max\left\{\, 1,  \, \Big\|  \frac{\bar{\w}-\w_i }{  \lambda/ \| \h^j\|_2^2 }\Big\|_2  \right\}}
,
\]
and the proximal step for the penalty term is
\[
\rmP^{1}_{\frac{1}{L_j}\frac{\gamma}{\sigma} \| \max\{-\cdot , \zeros \} \|_1}(\bar{\w}) 
~=~\textrm{median}\bigg( \bar{\w} + \frac{1}{L_j}\frac{\gamma}{\sigma}\ones  , \zeros, \bar{\w} \bigg)
~=~\textrm{median}\bigg( \bar{\w} + \frac{\gamma}{\| \h^j \|_2^2}\ones  , \zeros, \bar{\w} \bigg).
\]

Algorithm~\ref{algo21:prox_avg_W} performs one proximal-average iteration for 
$\textrm{update}(\W_k;\H_{k+1})$ in the BCD framework.
Repeating these iterations solve the W-subproblem \eqref{eqn:subproblemW}.
The per-iteration cost of the for-loop in Algorithm~\ref{algo21:prox_avg_W} is $\cO(r^2m)$, or $\cO(m^3)$ if $r \approx m$.

\begin{algorithm}[!h]
\DontPrintSemicolon
\For{$j=1,2,...,r$}{
Compute $\| \h^j \|_2^2$,  $\M_j = \M - \W\H + \w_j\h^j$\;
Update $\w_j$ by solving \eqref{eqn:subproblemwj} using one iteration of proximal-average as:

\qquad Compute $\bar{\w} = \M_j \h^j{^\top} / \| \h^j \|_2^2$
\;
\qquad  For $i \neq j$, compute\\
\qquad$
\rmP^{\frac{1}{L_j}\frac{\lambda}{\sigma}}_{\| \cdot - \w_i \|_2}(\bar{\w})
\,=\,
\rmP^{\frac{\lambda}{\| \h^j\|_2^2}}_{\| \cdot - \w_i \|_2}(\bar{\w})
\,=\,
\bar{\w} - \dfrac{\bar{\w} - \w_i}{\max\left\{\, 1,  \, \Big\|  \frac{\bar{\w}-\w_i}{  \lambda /  \| \h^j\|_2^2  }\Big\|_2  \right\}}
$
\;
\qquad Compute $\rmP^{1}_{\frac{1}{L_j}\frac{\gamma}{\sigma} \| \max\{-\cdot , \zeros \} \|_1}(\bar{\w})  = \textrm{median}\Big( \bar{\w} + \frac{\gamma}{\| \h^j \|_2^2}\ones  , \zeros, \bar{\w} \Big)$\;
\qquad $\displaystyle \w = \sum_{i \neq  j}^r \dfrac{\lambda}{\sigma} \rmP^{\frac{1}{L_j}\frac{\lambda}{\sigma}}_{\| \,\cdot\, - \w_i \|_2}(\bar{\w}) 
+ \dfrac{\gamma}{\sigma}\rmP^{1}_{\frac{1}{L_j}\frac{\gamma}{\sigma} \| \max\{-\cdot , \zeros \} \|_1}(\bar{\w}) 
$\;
}
\caption{A iteration of $\textrm{update}(\W_k;\H_{k+1},\M, \lambda,\gamma)$}
\label{algo21:prox_avg_W}
\end{algorithm}

\begin{remark}[Why not enforce $\W\geq \zeros$ as hard constraints?]
In NMF, nonnegativity is often imposed via an indicator function $\iota_{+}(\W)$, where $\iota_{+}(W_{ij}) = 0$ if $W_{ij} \geq 0$ and $\iota_{+}(W_{ij}) = +\infty $ otherwise. 
For SON-NMF, enforcing $\W \geq \zeros$ as a hard constraint may cause the proximal-average update to produce infeasible $\W$, resulting in the objective jumping to $+\infty$, and destroys the convergence of the whole method.
\end{remark}

\paragraph{Post-processing to extract columns of $\W$}
After minimizing the SON$_{2,1}$ norm with an overestimated rank, we select one representative column from each cluster to form the final rank-reduced matrix $\W$.

\section{Experiment}\label{sec:exp}
In this section we present numerical results to demonstrate the effectiveness of our algorithm for solving SON-NMF, and showcase SON-NMF's ability to identify the rank without prior knowledge.
In \cref{sec:exp:subsec:rank}, we evaluate SON-NMF's rank-revealing capability.
In \cref{sec:exp:subsec:time}, we compare the proposed algorithm's speed against ADMM and Nesterov's smoothing.
All the experiments were conducted on a Apple MacBook Air\footnote{M2 chipset, 8 CPU cores, 8 GPU cores with a 3.5GHz CPU and 8 GB RAM} in Python\footnote{Code is available:
\url{https://github.com/waqasbinhamed/sonnmf}}.

\subsection{SON-NMF identifies the rank without prior knowledge}\label{sec:exp:subsec:rank}
We test SON-NMF on datasets with known true NMF rank $r^*$.
The rank parameter $r$ is intentionally overestimated ($r > r^*$) to demonstrate that SON-NMF can correctly recover $r^*$.

\subsubsection{Synthetic data}\label{sec:exp:subsec:rank:syn}
We first use the synthetic dataset from \cite{leplat2019minimum}: $\Z =
\begin{bmatrix}
1 & 1 & 0 & 0
\\
0 & 0 & 1 & 1
\\
0 & 1 & 1 & 0
\\
1 & 0 & 0 & 1
\end{bmatrix}$
with $\rk(\Z) = 3 < 4 = \rk_{+}(\Z)$.

\paragraph{Dataset generation}
Let $\W_{\text{true}} = \Z$.
The ground truth $\H_{\text{true}}$ is generated by sampling each column from a Dirichlet distribution with parameter $\alpha = 0.05$.
The data matrix is then $\M = \W_{\text{true}}\H_{\text{true}} + \N$ 
where $\N \sim \cN(0,1)$ is Gaussian noise.

\paragraph{Experiment}
We solve \eqref{eqn:sonnmf} using inexact-BCD (Algorithm~\ref{algo:BCDframework}) with proximal average (Algorithm~\ref{algo21:prox_avg_W}) with:
\begin{itemize}[leftmargin=12pt]
    \item Random initialization of $\W$ and $\H$ in $[0,1)$.
    \item 1 update of $\H$ per 10 updates of $\W$.
    \item Stopping criterion: relative change $(F_k - F_{k-1}) / F_{k-1} < 10^{-6}$ or maximum iterations reached. 
    \item Table~\ref{table:param} shows the parameters used in the experiments.
\end{itemize}

\begin{table}[h!]
\centering
\caption{Parameters used in the algorithm in the experiments}
\begin{tabular}{lllll}
& $r$ & $\lambda$& $\gamma$ & max iteration  
\\ \hline\hline
synthetic data experiment 1 & $4$ & $10^{-6}$ & $10$ & $1000$
\\
synthetic data experiment 2 & $8$ & $10^{-6}$ & $1.5$ & $1000$
\\
swimmer & $50$ & $0.5$ & $10$ & $1000$
\\
Jasper experiment 1 & $64$ & $40000$  & $10000$ & $2000$
\\
Jasper experiment 2 & $100$ & $1000$ & $0.001$ & $1000$
\\
Jasper experiment 3 & $20$ & $1000000$& $1000000$ & $1000$
\\
Urban & $20$ & $1000000$ & $1000000$ & $1000$
\end{tabular}
\label{table:param}
\end{table}   

\begin{figure}[h!]
\centering\includegraphics[width=\textwidth]{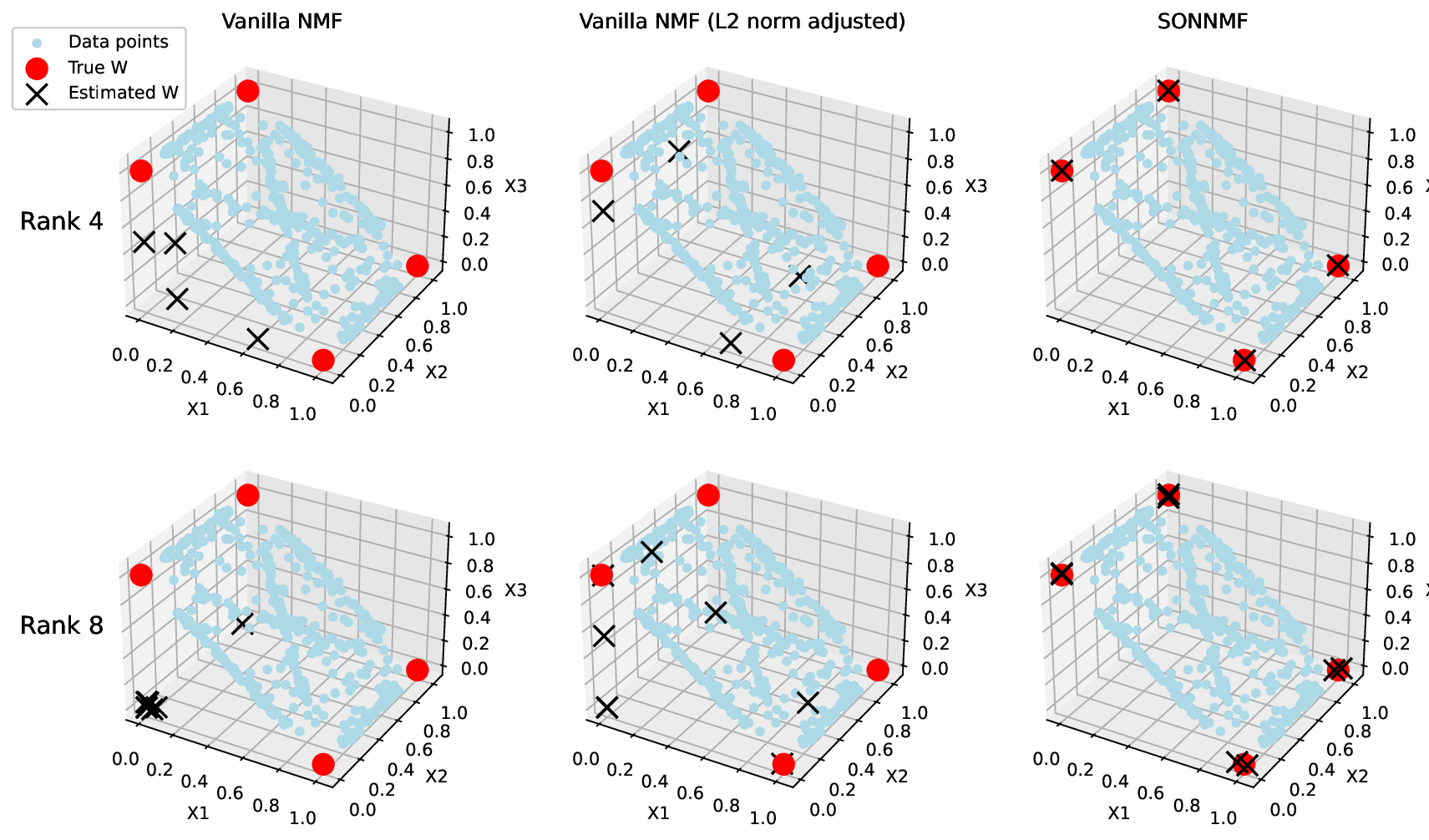}
\caption{
Reconstructed columns of $\W$ (crosses) by NMF and SON-NMF, compared with the ground truth columns (red dots).
\textbf{Left}: $\W$ from NMF; \textbf{Middle}: NMF with columns normalized to 1; \textbf{Right}: $\W$ from SON-NMF.
In both cases $r=4$ and $r=8$, the crosses given by SON-NMF fit numerically with the red dots.
}
\label{fig:syn}
\end{figure}

\begin{figure}[h!]
\centering
\includegraphics[width=0.5\textwidth]{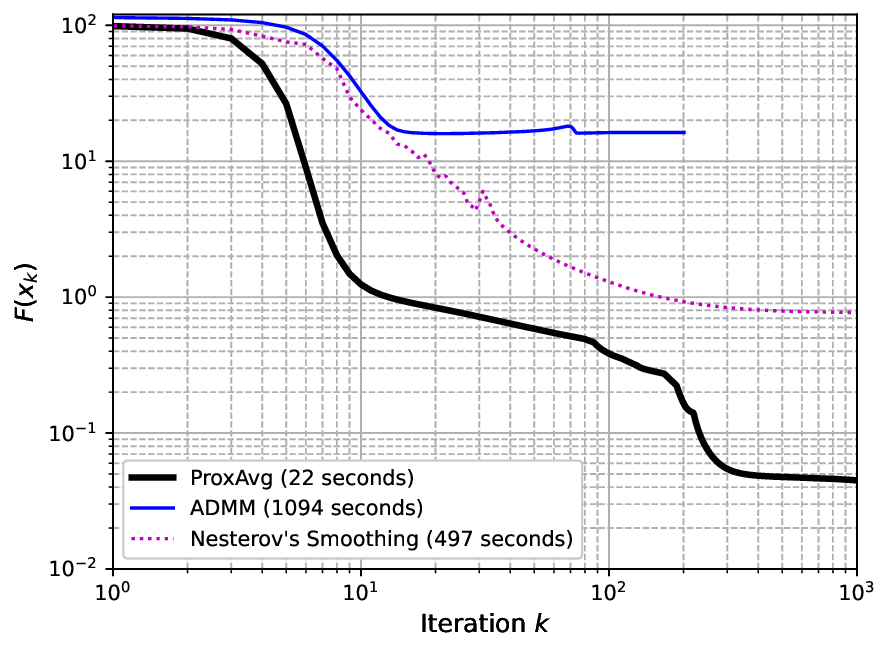}
\includegraphics[width=0.49\textwidth]{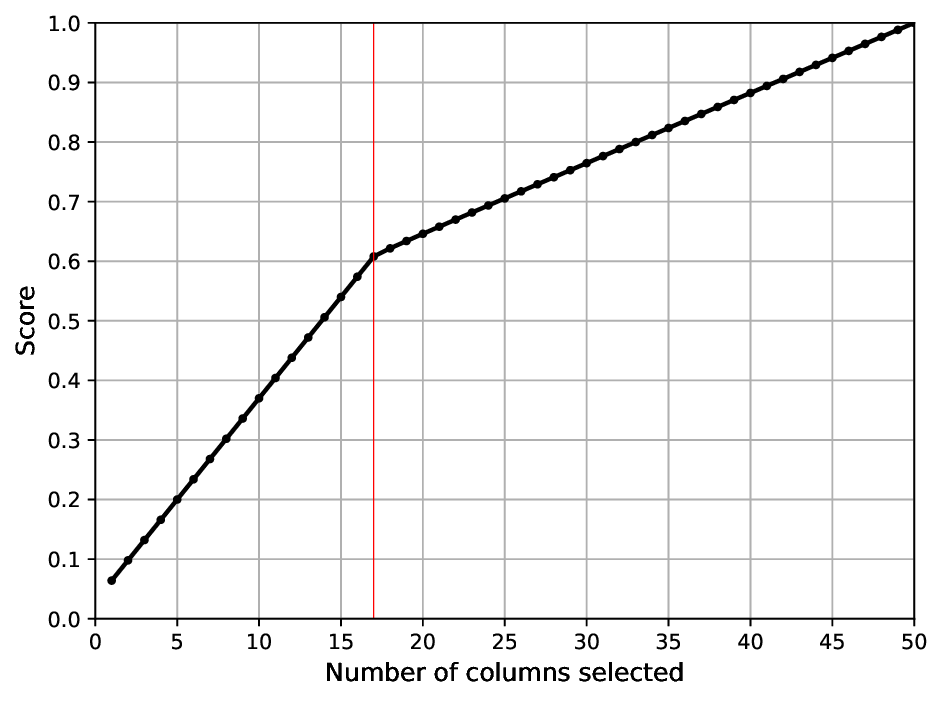}
\caption{
\textbf{Left}: 
Convergence of the SON-NMF cost function on synthetic data (Experiment 2).
We compare three BCD algorithms for solving the $W$-subproblem: proximal average (this work), ADMM, and Nesterov's smoothing.
Computation time (in seconds) is also shown. Proximal average achieves the fastest convergence.
\textbf{Right}: Column-selection score (SON term) on the swimmer dataset using a simple greedy search. The red line at $r=17$ marks the cut-off, matching the true number of components in the dataset.
}
\label{fig:F_clipped_sythetic_r8}
\end{figure}

\paragraph{Result}
Fig.~\ref{fig:syn} shows that SON-NMF reconstructs the data more accurately than standard NMF.
Fig.~\ref{fig:F_clipped_sythetic_r8} compares the convergence of W-subproblem updates using BCD with proximal averaging, BCD with ADMM, and BCD with Nesterov's smoothing, showing faster convergence for the proposed method.

\subsubsection{The swimmer dataset}
\begin{figure}[h!]\centering
\includegraphics[width=\textwidth]{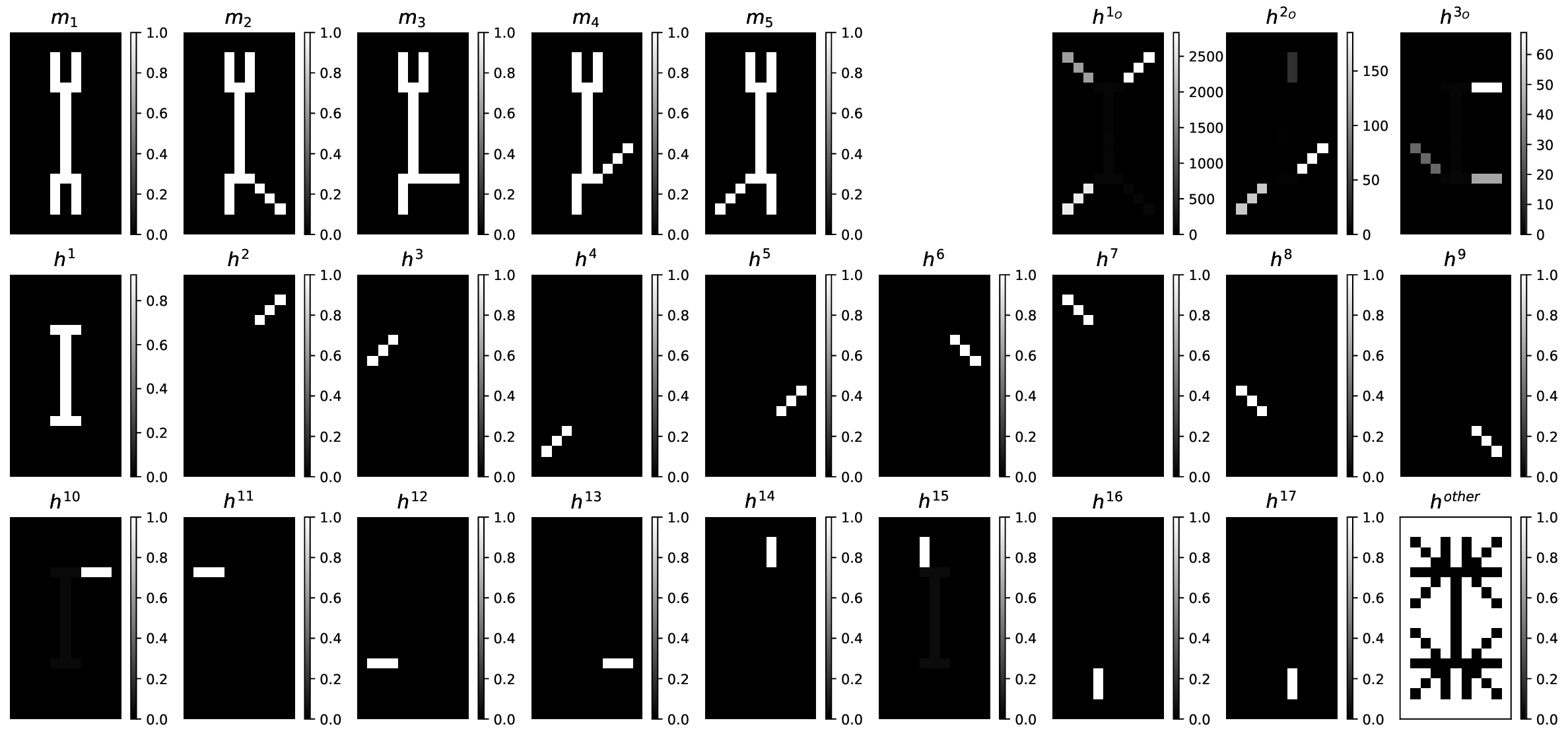}
\caption{Top row (left): first 5 images ($\m^1,\m^2,\dots,\m^5$) in the swimmer dataset, showing a swimmer swimmer in motion.
Top row (right): 3 $\h^j_{\text{O}}$ obtained from rank-50 vanilla NMF (subscript O denotes the vanilla NMF). 
These components exhibit mixed or overlapping factors.
Bottom rows: Decomposition from rank-50 SON-NMF.
Here $\h^1$ captures the torso, $\h^2, \h^3,\dots,\h^{17}$ capture the four limbs.
The remaining components are aggregated into $\h^{\text{other}}$, which represents noise.
$\h^{\text{other}}$ is complementary to all $\h^1,\dots, \h^{17}$, and its corresponding $\w$ has negligible energy (not plotted).
}
\label{fig:swimmer}
\end{figure}

We next use the \texttt{swimmer} dataset\footnote{\url{https://gitlab.com/ngillis/nmfbook/}} introduced by \cite{donoho2003does}.
It consists of 256 images of size $20 \times 11$ pixels representing a skeleton ``swimming'' (top row of Fig.~\ref{fig:swimmer}).
By inspection, the true NMF rank is $r^* = 17$: 1 for the torso and 16 for the four limbs (4 movements per limb).
We apply SON-NMF with an overestimated rank $r = 50 > r^*$. 
All 17 true components are successfully recovered, while the extra components capture small-energy noise.
Using a simple greedy search to select columns of $\W$, the score plot (right of Fig.~\ref{fig:F_clipped_sythetic_r8}) shows a clear cut-off at $r = 17$, accurately identifying the true rank.

\subsubsection{Jasper ridge hyperspectral dataset}\label{sec:exp:subsec:rank:jasper}
\begin{figure}[h!]
    \centering
    \includegraphics[width=0.33\textwidth]{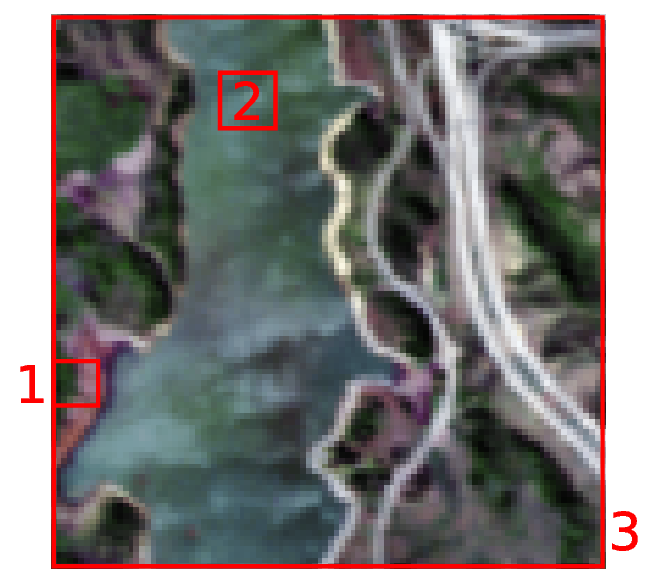}
    \includegraphics[width=0.59\textwidth]{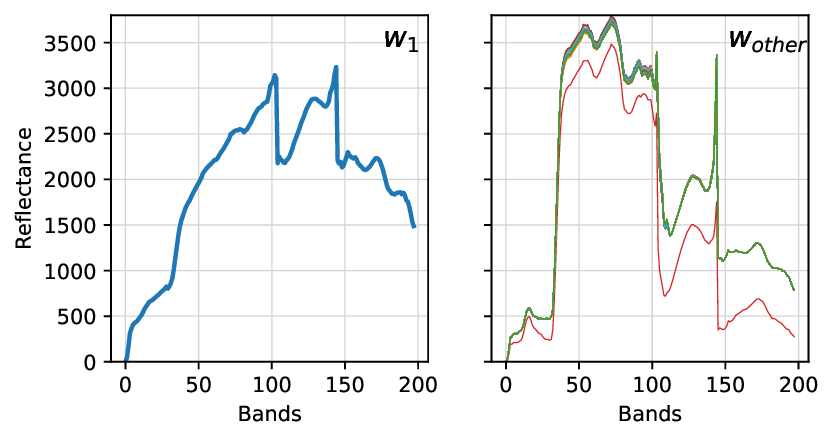}
    \caption{\textbf{Left}: Photograph of the Jasper Ridge site with the three selected regions highlighted in red.  
    \textbf{Right}: Result for Jasper experiment 1.
    SON-NMF separates two distinct materials: soil (captured by $\w_{1}$) and vegetation (captured collectively by $\w_{\text{other}}$, i.e., all remaining columns of $\W$).
    }
    \label{fig:region}
\end{figure}

We next evaluate SON-NMF on the Jasper Ridge hyperspectral dataset\footnote{Available in MATLAB: \url{https://uk.mathworks.com/help/images/explore-hyperspectral-data-in-the-hyperspectral-viewer.html}}
This dataset has dimensions $100 \times 100 \times 198$, corresponding to $100 \times 100$ spatial pixels and 198 spectral bands (wavelength channels).
Background on applying NMF to hyperspectral images can be found in \cite[Sect.~1.3.2]{gillis2020nonnegative}.
Fig.~\ref{fig:region} shows a photograph of the Jasper Ridge site along with the three spatial regions selected for our experiments.
Because the dataset entries have relatively large numerical values, we scale the SON regularization parameter $\lambda$ to a higher magnitude (see Table~\ref{table:param}).

\paragraph{Jasper experiment 1}
We apply rank-$64$ SON-NMF to an $8 \times 8$ region containing vegetation and soil.  
Here we set $r = 64 = mn$, i.e., the rank is as large as the size of the dataset.  
Fig.~\ref{fig:region} shows the matrix $\W$ obtained from SON-NMF.  
By inspection, region~1 consists of two end-member materials: soil and vegetation.  
SON-NMF successfully identifies these two materials, confirming its ability to recover the correct number of components without prior knowledge of the true rank.

\paragraph{Jasper experiment 2}
We apply rank-$100$ SON-NMF to a $10 \times 10$ region consisting purely of water.  
Since this region contains only one material, the correct decomposition rank is $1$.  
SON-NMF successfully recovers the water component, effectively reducing a rank-$100$ initialization to a rank-$1$ solution.  

We remark that in this special case, the rank-$1$ factorization can also be obtained algebraically.  
By the Perron--Frobenius theorem, the leading component in the eigendecomposition of the covariance matrix yields the exact rank-$1$ NMF solution (see Proposition~\ref{prop:pf-rank1}).

\begin{figure}[h!]
    \centering
    \includegraphics[width=\textwidth]{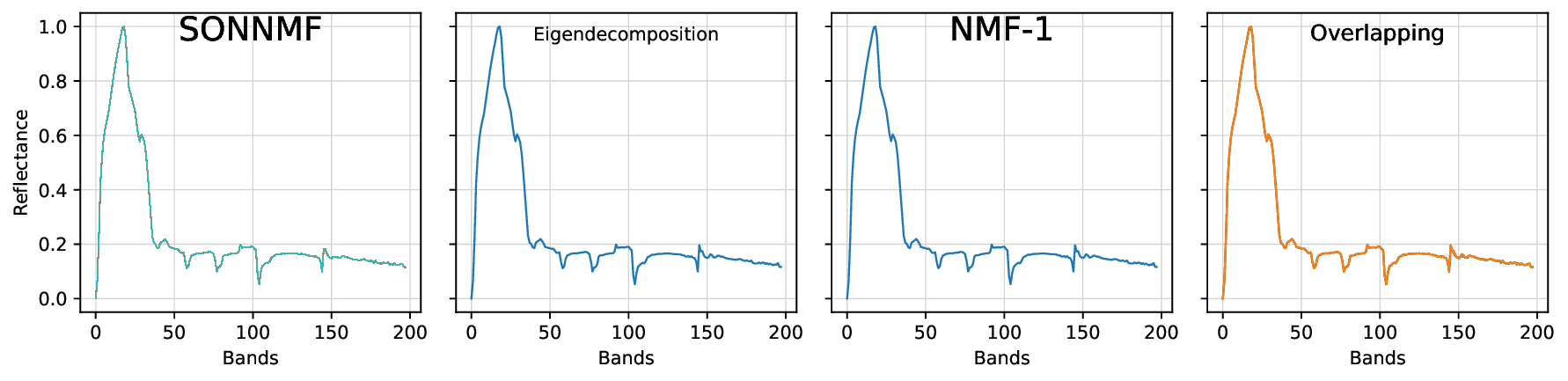}
    \caption{
    Result for Jasper experiment 2.
    The rank-100 SON-NMF (with $r=100$, much larger than the true rank $r^*=1$) identifies the water spectrum.  
    \textbf{Left:} All 100 columns of $\W$ from SON-NMF share the same waveform.  
    \textbf{Middle:} $\W$ obtained from eigendecomposition and rank-1 vanilla NMF.  
    \textbf{Right:} Overlapping plot of all $\W$ columns, showing SON-NMF agrees with the vanilla NMF solution.  
    All columns are normalized to unit $\ell_\infty$-norm for clarity.
    }
    \label{fig:exp2}
\end{figure}

\begin{proposition}\label{prop:pf-rank1}
Given a data matrix $\M \in \IRmn_+$ with NMF $\M = \W\H$, and columns of $\W$ ordered according to $\|\w_j \h^j\|$, then for $r=1$ (rank-1 NMF), the leading column $\w_1$ of $\W$ is given by the leading eigenvector of $\M \M^\top$.
\begin{proof}
$\M\M^\top = \W\H\H^\top\W^\top = \W \G \W^\top$ with $\G := \H\H^\top$.  
Let $\G = \V \bSigma \V^\top$ and $\M\M^\top = \U \bLambda \U^\top$ be the eigendecompositions. Then $\W \V \bSigma \V^\top \W^\top = \U \bLambda \U^\top$ implies
\[
\W\V = \U \implies (\W\V)_{:,1} = \U_{:,1} \iff \W \v_1 = \u_1.
\]
Both $\G = \H\H^\top$ and $\M\M^\top$ are nonnegative square matrices.
By the Perron--Frobenius theorem, both $\u_1$ and $\v_1$ are nonnegative. Thus $\u_1 \in \cone(\W)$, and for $\rk(\W) = 1$, we have $\u_1 = \w_1$.
\end{proof}
\end{proposition}
Fig.~\ref{fig:exp2} shows that SON-NMF recovers the water spectrum with a relative error of $0.006$, matching the exact eigendecomposition solution.

\begin{figure}[h!]
    \centering
    \includegraphics[width=\textwidth]{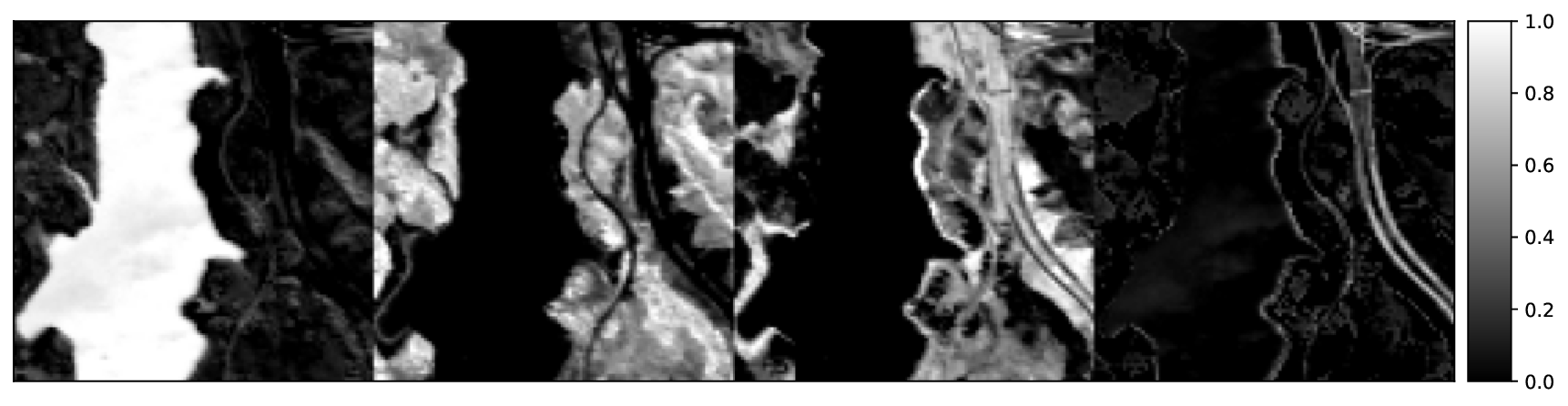}
    \includegraphics[width=\textwidth]{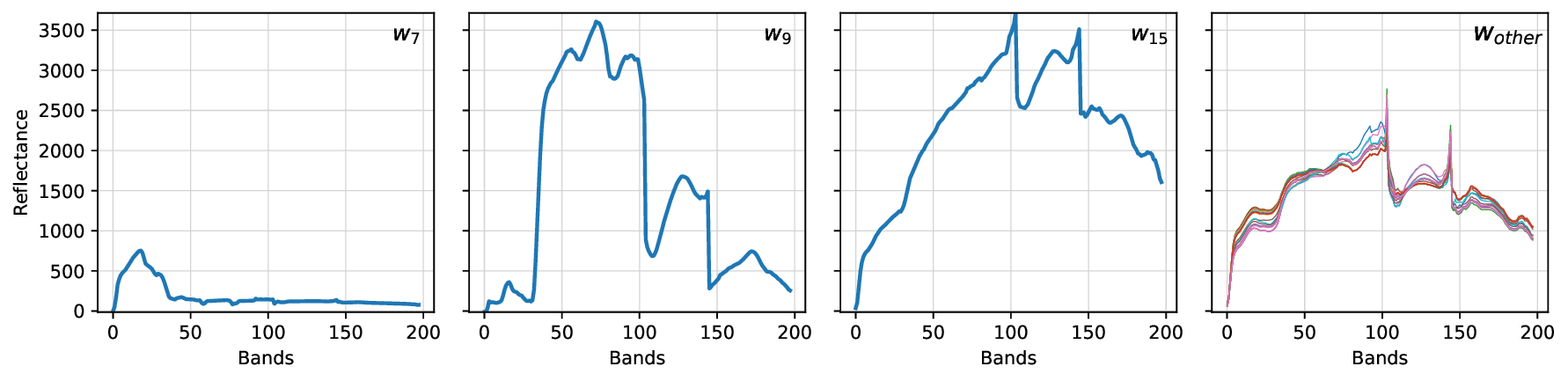}
    \caption{Result for Jasper experiment 3.
    Four material are extracted: (from left to right) water, vegetation, soil and road.
    }
    \label{fig:exp3}
\end{figure}

\paragraph{Jasper experiment 3}
n this experiment, we run a rank-20 SON-NMF on the full Jasper Ridge dataset.  
SON-NMF extracts four distinct materials, as shown in Fig.~\ref{fig:exp3}.  
The extracted materials correspond to water, vegetation, soil, and road, and agree well with results obtained from other hyperspectral unmixing methods.

\subsubsection{The Urban hyperspectral dataset}\label{sec:exp:subsec:rank:urban}
\begin{figure}[h!]
\centering
\includegraphics[width=\textwidth]{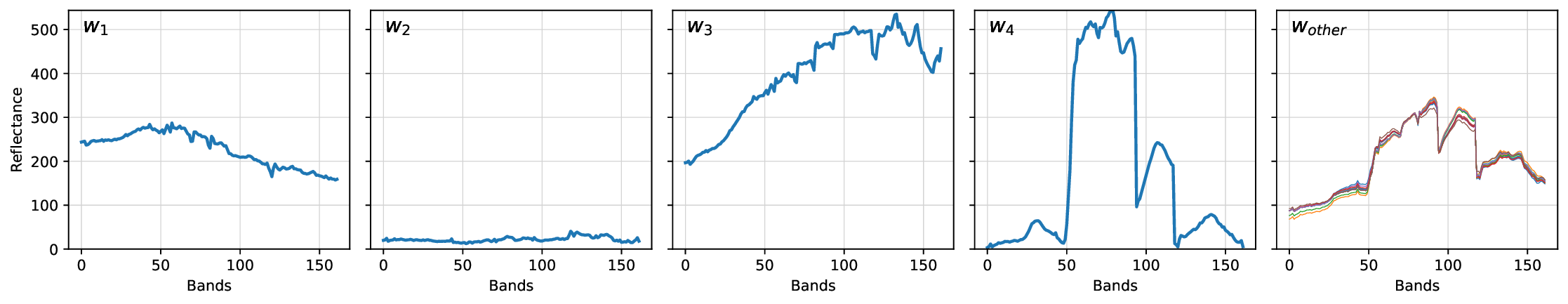}    
\includegraphics[width=0.17\textwidth]{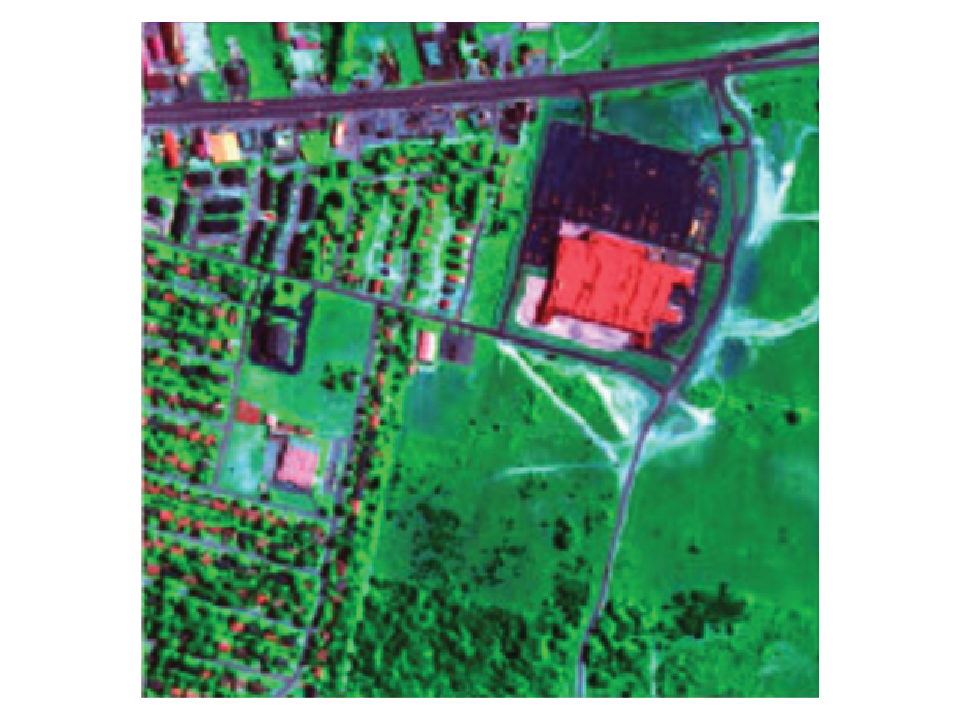}   
\includegraphics[width=0.82\textwidth]{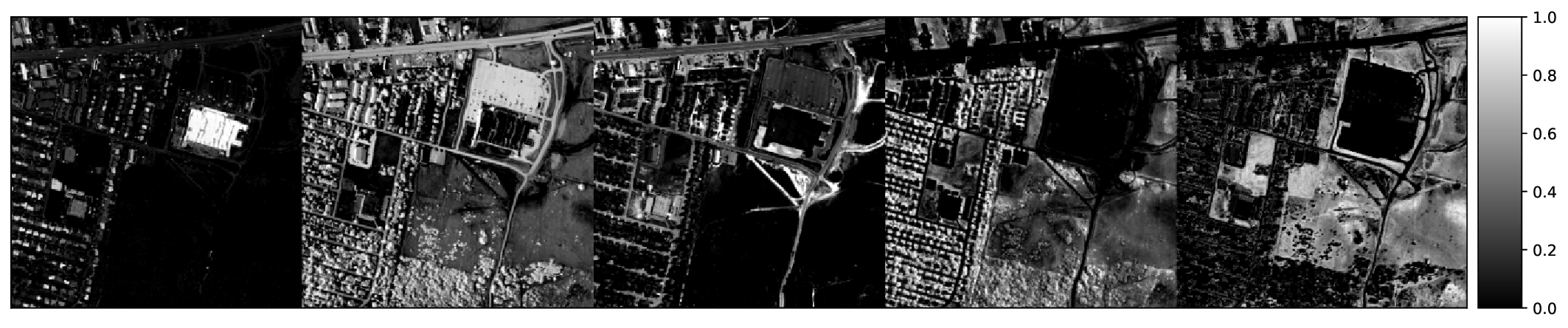}  
\caption{
SON-NMF decomposition of the Urban hyperspectral dataset using rank-20.  
\textbf{Top:} Overlapped spectral signatures (columns of $\W$) of the five extracted materials: roof, asphalt, soil, tree, and grass (from left to right).  
\textbf{Bottom left:} The true photo of the Urban dataset.  
\textbf{Bottom right:} Summed abundances ($\H$) of the extracted materials across the spatial domain.  
Note that the weak component, asphalt, is successfully extracted by SON-NMF, which is not the case with classical NMF or other rank estimation approaches.
}
\label{fig:urban_result}
\end{figure}

In this section, we conduct an experiment on a large-scale dataset with approximately $1.5 \times 10^7$ data points.  
We use the Urban dataset\footnote{Available at \url{https://gitlab.com/ngillis/nmfbook/}}, which is a $307 \times 307 \times 162$ data cube, with pixel dimensions $307 \times 307$ and 162 spectral bands.  
We run a rank-20 SON-NMF on this dataset with parameters $\lambda = \gamma = 10^6$, allowing up to 1000 iterations.  
SON-NMF successfully identifies five distinct material clusters, see Fig.\ref{fig:urban_result}.

\subsection{Speed of the algorithm}\label{sec:exp:subsec:time}
Fig.~\ref{fig:F_clipped_sythetic_r8} shows the convergence of BCD (Algorithm~\ref{algo:BCDframework}) using proximal average to solve the W-subproblem (Algorithm~\ref{algo21:prox_avg_W}), compared with BCD with ADMM and BCD with Nesterov's smoothing.  
The results clearly indicate that proximal average outperforms the other methods.  
For a detailed discussion on why proximal average performs better than smoothing, see \cite{yu2013better}.

\paragraph{ADMM is not suitable for SON-NMF: high per-iteration cost}
The subproblem \eqref{min:phipsi} of size $n \times 1$ can be solved using multi-block ADMM, which introduces $N$ auxiliary variables and $N$ Lagrange multipliers.  
The resulting augmented Lagrangian has size $n \times (1+2N)$, leading to a large computational overhead.  
Specifically, the mapping $\W \mapsto P(\W)$ is $m \times r$ to $m \times r(r-1)/2$ due to the numerous nonsmooth terms $\|\w_i - \w_j\|_2$ in the SON regularization.  
For each column $\w_i$, the number of nonsmooth terms is $r$, making the per-iteration complexity for multi-block ADMM $\cO(2 m r^2 + m r)$.  
In contrast, proximal average has per-iteration complexity $\cO(m r)$.  
When $r \sim m$, this means ADMM requires $\cO(2 m^3 + m^2)$ per iteration, whereas proximal average only requires $\cO(m^2)$.  
Combined with the generally slower convergence of ADMM, this makes it inefficient for solving the W-subproblem in SON-NMF.

\subsection{Discussion: favourable features of SON-NMF for applications}\label{sec:app}
We summarize the key advantages of SON-NMF observed in our experiments.

\paragraph{Empirically rank-revealing and handles rank deficiency}
All seven experiments in \cref{sec:exp} demonstrate that SON-NMF can learn the factorization rank without prior knowledge.
SON-NMF is effective on datasets with rank deficiency, a feature not present in other regularized NMF models such as minvol NMF \cite{leplat2019minimum}, despite their empirical rank-revealing capabilities.

\paragraph{Detects weak components}
The clustering property of the SON term allows SON-NMF to identify weak components that vanilla NMF often misses. 
For example, in the Jasper dataset (\cref{sec:exp}), the water component contributes only $9\%$ of the total energy $\| \w_{\text{water}} \h^{\text{water}}\|_F / \| \M \|_F$, compared with $54\%$ for vegetation.  
The squared Frobenius norm in vanilla NMF emphasizes high-energy components, potentially ignoring weaker ones.  
In contrast, SON-NMF, through the term $\| \w_{\text{other}} - \w_{\text{water}}\|_2$, successfully extracts the water component.  
Lemma~\ref{lem:SON20min} further guarantees that the smallest possible cluster identified by the SON term has size at least 1.

\paragraph{Handles spectral variability}
The $\W$ in hyperspectral experiments (Figs.~\ref{fig:region},~\ref{fig:exp2},~\ref{fig:exp3},~\ref{fig:urban_result}) naturally capture spectral variability \cite{borsoi2021spectral}.  
Thus, SON-NMF can effectively model such variability without requiring complex preprocessing pipelines.

\begin{figure}[h!]\label{fig:jasper_diff_range}
\centering\includegraphics[width=\textwidth]{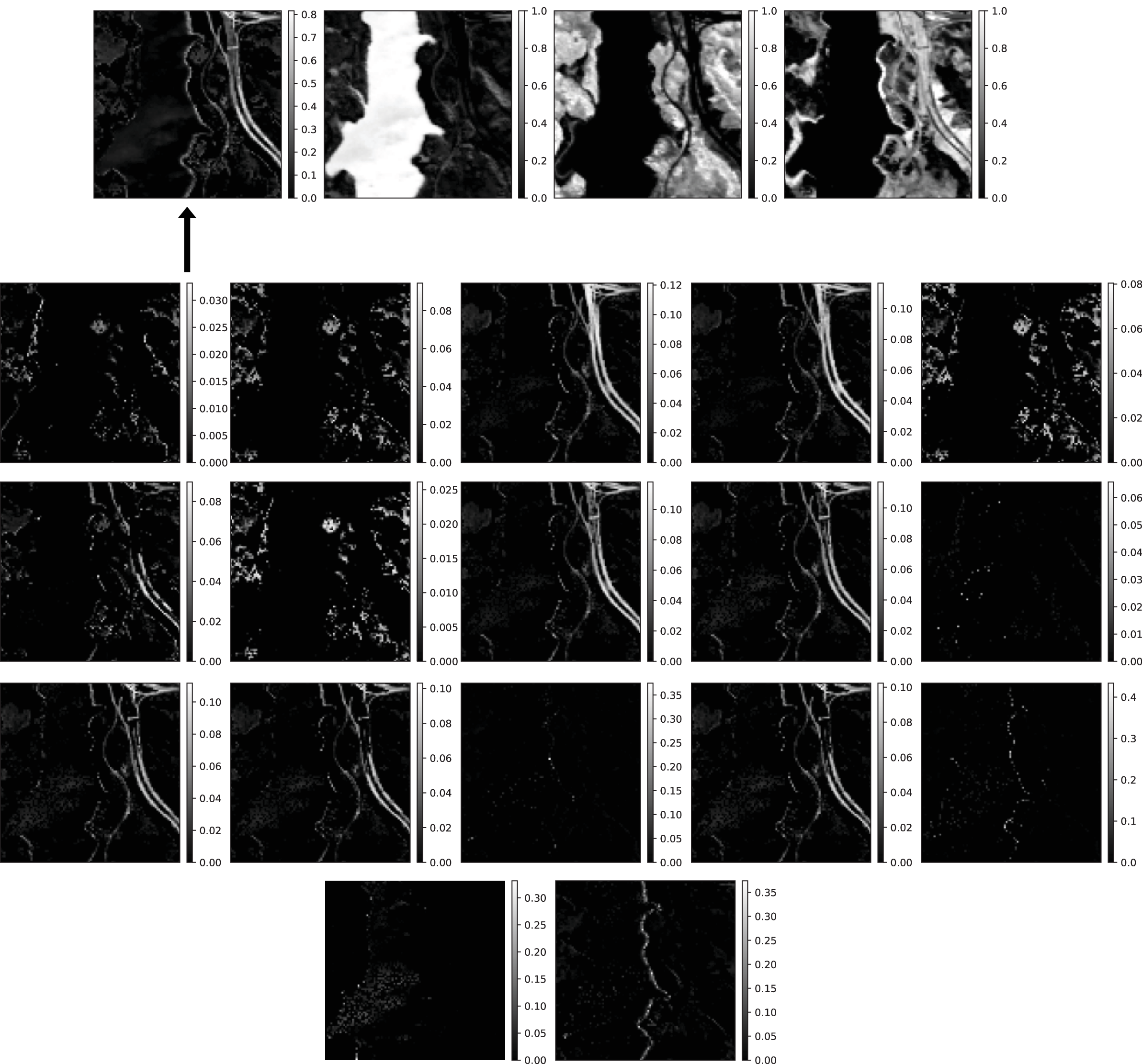}
\caption{Full decomposition map of SON-NMF ($r=20$) on Jasper dataset ($r^* \approx 4$).  
Here the road endmember consists of 17 components.}
\end{figure}

\paragraph{Hierarchical clustering capability}
The number of clusters obtained by SON-NMF depends on the regularization parameter $\lambda$.  
Smaller $\lambda$ values yield finer clusters, while larger values produce coarser clusters.  
This hierarchical nature is advantageous for datasets with multi-scale structure, as demonstrated in hyperspectral experiments, see Figs.~\ref{fig:jasper_diff_range}.

\section{Conclusion}\label{sec:conclusion}
In this paper, we proposed a sum-of-norm (SON) regularized NMF model, designed to estimate the factorization rank in NMF on-the-fly.  
The resulting SON-NMF problem is nonconvex, nonsmooth, non-separable, and non-proximal. 
To solve it, we developed a block coordinate descent (BCD) algorithm combined with proximal averaging.  

Theoretically, we showed that the complexity of the SON term in SON-NMF is irreducible, implying that the computational cost of solving SON-NMF can be very high.  
This is expected, as estimating the rank in NMF is an NP-hard problem.  

Empirically, we demonstrated that SON-NMF can accurately detect the correct factorization rank and extract weak components, making it particularly suitable for applications in imaging and hyperspectral data analysis.  
Its hierarchical clustering property and ability to handle spectral variability further highlight its practical advantages.

\section*{Acknowledgement}
Andersen Ang thanks Steve Vavasis for the discussion on graph theory and the complexity of SON-NMF.

\bibliographystyle{apalike}
\bibliography{refs}

\end{document}